\documentclass{article}

\usepackage{microtype}
\usepackage{graphicx}
\usepackage{subfigure}
\usepackage{booktabs} 

\usepackage{hyperref}

\usepackage[accepted]{icml2024}

\usepackage{amsmath}
\usepackage{amssymb}
\usepackage{mathtools}
\usepackage{amsthm}
\usepackage{amsfonts}
\usepackage{multirow}
\usepackage{bbm}
\usepackage[capitalize,noabbrev]{cleveref}

\theoremstyle{plain}

\theoremstyle{definition}

\theoremstyle{remark}

\usepackage[textsize=tiny]{todonotes}

\icmltitlerunning{Neighboring Perturbations of Knowledge Editing on Large Language Models}

\newcommand{\MODELNAME}{APP}

\begin{document}

\twocolumn[
\icmltitle{Neighboring Perturbations of Knowledge Editing on Large Language Models}




\icmlsetsymbol{equal}{*}

\begin{icmlauthorlist}
\icmlauthor{Jun-Yu Ma}{ustc}
\icmlauthor{Zhen-Hua Ling}{ustc}
\icmlauthor{Ningyu Zhang}{zju}
\icmlauthor{Jia-Chen Gu}{ucla}
\end{icmlauthorlist}

\icmlaffiliation{ustc}{NERC-SLIP, University of Science
and Technology of China}
\icmlaffiliation{ucla}{University of California, Los Angeles}
\icmlaffiliation{zju}{Zhejiang University}

\icmlcorrespondingauthor{Jia-Chen Gu}{gujc@ucla.edu}

\icmlkeywords{Machine Learning, ICML}

\vskip 0.3in
]



\printAffiliationsAndNotice{}  

\begin{abstract}
    Despite their exceptional capabilities, large language models (LLMs) are prone to generating unintended text due to false or outdated knowledge. Given the resource-intensive nature of retraining LLMs, there has been a notable increase in the development of \emph{knowledge editing}. 
    However, current approaches and evaluations rarely explore the perturbation of editing on neighboring knowledge.
    This paper studies whether updating new knowledge to LLMs perturbs the neighboring knowledge encapsulated within them.
    Specifically, we seek to figure out whether appending a new answer into an answer list to a factual question leads to catastrophic forgetting of original correct answers in this list, as well as unintentional inclusion of incorrect answers.
    A metric of \emph{additivity} is introduced and a benchmark dubbed as Perturbation Evaluation of Appending Knowledge (PEAK) is constructed to evaluate the degree of perturbation to neighboring knowledge when appending new knowledge. 
    Besides, a plug-and-play framework termed Appending via Preservation and Prevention (\MODELNAME{}) is proposed to mitigate the neighboring perturbation by maintaining the integrity of the answer list.
    Experiments demonstrate the effectiveness of \MODELNAME{} coupling with four editing methods on four LLMs. 
    The code and data are available at \href{https://github.com/mjy1111/PEAK}{https://github.com/mjy1111/PEAK}.
\end{abstract}

\begin{figure}[ht]
\centering
\includegraphics[width=0.48\textwidth]{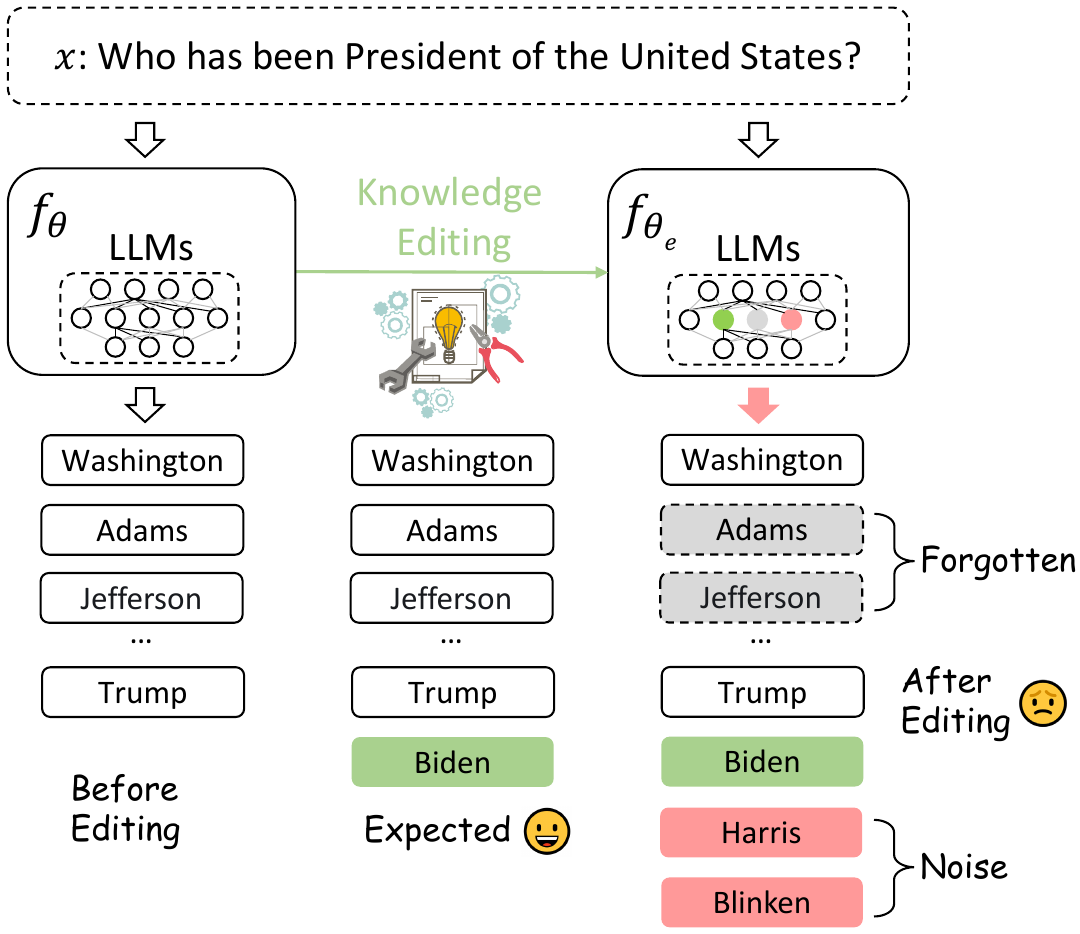}
\vspace{-6mm}
\caption{
Illustration of the neighboring perturbations while appending a new answer into an answer list to a factual question.
Catastrophic forgetting of original correct answers and unintentional inclusion of incorrect answer are both undesirable.
$f_{\theta}$ / $f_{\theta_{e}}$ denotes the models before / after editing.} \label{figure1}
\vspace{-4mm}
\end{figure}

\vspace{-4mm}
\section{Introduction}
Large language models (LLMs) such as GPT-4~\cite{DBLP:journals/corr/abs-2303-08774} have demonstrated remarkable capabilities in Natural Language Processing (NLP)~\cite{DBLP:conf/nips/Ouyang0JAWMZASR22,DBLP:conf/icml/0006HB23, DBLP:journals/corr/abs-2307-09288}.
Nonetheless, current LLMs often inevitably exhibit hallucinations stemming from outdated or erroneous knowledge within their parameters~\cite{DBLP:journals/corr/abs-2309-01219, DBLP:journals/corr/abs-2302-12813, DBLP:journals/csur/JiLFYSXIBMF23,DBLP:journals/corr/abs-2311-05232}. 
Given that retraining LLMs is both costly and time-consuming, 
there has been a surge in research focused on \emph{knowledge editing} (a.k.a., \emph{model editing})~\cite{DBLP:conf/iclr/SinitsinPPPB20,DBLP:journals/corr/abs-2012-00363,DBLP:conf/emnlp/CaoAT21, DBLP:conf/nips/MengBAB22, DBLP:conf/iclr/MengSABB23, DBLP:conf/acl/DaiDHSCW22,DBLP:conf/icml/MitchellLBMF22,DBLP:conf/emnlp/YaoWT0LDC023,DBLP:conf/emnlp/ZhongWMPC23, DBLP:journals/corr/abs-2307-12976, DBLP:journals/corr/abs-2310-10322,DBLP:journals/corr/abs-2401-04700}, which aims at efficiently altering LLMs’ behaviors within specific domains while preserving overall performance across various inputs.

Many researchers endeavor to develop editing
methods to modify the parameters of models, which can be generally classified into two categories of \emph{Meta-Learning} and \emph{Locate-then-Edit}~\cite{DBLP:journals/corr/abs-2308-07269, DBLP:conf/emnlp/YaoWT0LDC023, zhang2024comprehensive}.
Numerous benchmarks have been designed to assess these methods across various dimensions, the most fundamental of which are \emph{Efficacy}, \emph{Generalization}, and \emph{Locality}~\cite{DBLP:conf/nips/MengBAB22,DBLP:conf/emnlp/ZhongWMPC23,DBLP:journals/corr/abs-2307-12976,DBLP:journals/corr/abs-2310-10322}. 
These approaches and benchmarks primarily focus on determining if the new target knowledge has been successfully memorized.
However, the perturbations of editing on knowledge neighboring to the new target knowledge have not been fully explored when updating new knowledge to LLMs.

In this study, we particularly investigate whether the editing operation of appending a new answer into an answer list to a factual question perturbs the neighboring knowledge encapsulated within them, as illustrated in Figure~\ref{figure1}. 
Specifically, we seek to figure out whether the knowledge appending leads to catastrophic forgetting of original correct answers in this answer list, as well as the unintentional inclusion of incorrect answers.
In addition to the metrics that are commonly used in previous works, a new metric of \emph{additivity} is introduced to assess the degree of
perturbation to neighboring knowledge when appending.
To evaluate the additivity of edited models, a benchmark dubbed as \textbf{P}erturbation \textbf{E}valuation of \textbf{A}ppending \textbf{K}nowledge (PEAK) is constructed, with each example comprising a factual question, a list of original correct answers, a list of incorrect distractors, and a piece of knowledge to append.
After an editing operation of knowledge appending, the list of original correct answers is utilized to calculate the proportion of them that are accidentally removed.
Similarly, the list of incorrect distractors is utilized to calculate the proportion of them that are unintentionally included.
PEAK includes two datasets of PEAK-CF based on counterfactual edits, and of PEAK-T based on temporal knowledge edits of changes in real-world.

To mitigate the neighboring perturbations after appending new knowledge, a plug-and-play framework named \textbf{A}ppending via \textbf{P}reservation and \textbf{P}revention (\MODELNAME{}) is proposed.
A set of editing objectives is designed to minimize the probability perturbations of both correct and incorrect knowledge. 
On the one hand, a certain margin is maintained between the probability of correct knowledge and that of incorrect knowledge. 
On the other hand, it involves ensuring that the probability of correct knowledge does not decrease while controlling that the probability of incorrect knowledge does not increase.
In this way, this framework helps preserve the integrity of original correct knowledge and prevent the inclusion of false knowledge while appending new knowledge.
Furthermore, the proposed editing framework eliminates the need for a training step, and is adaptable to be coupled with multiple existing editing methods.

To investigate the performance of current knowledge editing methods for appending knowledge, this study conducts comprehensive experiments that encompass a wide range of methods, including FT~\cite{DBLP:journals/corr/abs-2012-00363}, MEND~\cite{DBLP:conf/iclr/MitchellLBFM22}, KN~\cite{DBLP:conf/acl/DaiDHSCW22}, ROME~\cite{DBLP:conf/nips/MengBAB22} and MEMIT~\cite{DBLP:conf/iclr/MengSABB23}. 
Four representative LLMs of varying sizes are utilized as our foundational models, including GPT-2 XL (1.5B)~\cite{radford2019language}, GPT-J (6B)~\cite{wang2021gpt}, LLaMA-2 (7B) and LLaMA-2 (13B)~\cite{DBLP:journals/corr/abs-2307-09288}.  
We surprisingly observe that although current editing methods can effectively incorporate new facts, they significantly undermine the probability distribution of knowledge neighboring to the new facts, disrupting the integrity of original correct knowledge and introducing unintentional noise.
Furthermore, experimental results extensively demonstrate the effectiveness of the proposed \MODELNAME{} in mitigating the neighboring perturbations of editing, as well as its compatibility with four editing methods on four representative LLMs of different sizes.

In essence, our research offers three significant contributions:
(1) This study pioneers the exploration of neighboring perturbations via appending new knowledge to LLMs.
A metric of additivity is introduced and a new benchmark PEAK is constructed to gauge the degree of perturbation to neighboring knowledge. 
(2) Through comprehensive experiments, we observe that although established methods and LLMs absorb new knowledge effectively, 
they seriously disrupt the integrity of original correct knowledge and introduce unintentional false knowledge.
(3) The plug-and-play \MODELNAME{} is proposed which is effective in mitigating neighboring perturbations when appending knowledge.
We aspire that both our benchmark and method can shed light on the neighboring perturbations of knowledge editing on LLMs.

\section{Related Work}

Many knowledge editing methods have been proposed to modify knowledge encoded in model, such as meta-learning, and locate-then-edit \cite{DBLP:conf/emnlp/YaoWT0LDC023}.
On the one hand, meta-learning methods train a hypernetwork to get gradient changes to update model parameters~\cite{DBLP:conf/emnlp/CaoAT21,DBLP:conf/iclr/MitchellLBFM22}.
MEND~\cite{DBLP:conf/iclr/MitchellLBFM22} learns to transform the fine-tuning gradient into a low-rank decomposition of the gradient.
On the other hand, locate-then-edit methods first locate knowledge neurons in LLMs that exhibit a positive correlation with a knowledge expression, and then modify them accordingly~\cite{DBLP:conf/acl/DaiDHSCW22,DBLP:conf/nips/MengBAB22,DBLP:conf/iclr/MengSABB23}.
\citet{DBLP:conf/acl/DaiDHSCW22} computed the contribution of each neurons to a certain knowledge, then updated or erased knowledge by modifying these neurons with
the embedding vectors of facts.
\citet{DBLP:conf/nips/MengBAB22} located multi-layer perceptron (MLP) storing factual knowledge, and then edited such knowledge by injecting new key-value pair in the MLP module, which follows recent observations that these layers can
be cast as key-value memories that store factual knowledge~\cite{DBLP:conf/emnlp/GevaSBL21,DBLP:conf/emnlp/GevaCWG22}.
In addition, \citet{DBLP:journals/corr/abs-2012-00363} also propose the constrained fine-tuning approach on modified facts.
DeepEdit~\cite{wang2024deepedit} is neuro-symbolic that decodes with constraints and can be flexibly applied to black-box LLMs.
Besides, some benchmarks are proposed for assessing knowledge editing~\cite{DBLP:conf/nips/MengBAB22, DBLP:journals/corr/abs-2307-12976, DBLP:journals/corr/abs-2310-10322, DBLP:journals/corr/abs-2310-02129,DBLP:journals/corr/abs-2303-07345}.
For example, MQUAKE~\cite{DBLP:conf/emnlp/ZhongWMPC23} uses the multi-hop questions to assess knowledge editing, suggesting that editing a particular fact implies that many other facts need to be updated.
BAKE~\cite{DBLP:journals/corr/abs-2310-10322} assesses knowledge editing bidirectionally.
\citet{DBLP:journals/corr/abs-2310-02129} have explored two significant areas of concern: Knowledge Conflict and Knowledge Distortion, which aims to unveil more weakness in knowledge editing.
Additionally, editing has been applied in various domains, such as changing model personality~\cite{DBLP:journals/corr/abs-2310-02168}, editing multimodal models~\cite{DBLP:conf/emnlp/0008TL0WC023}, protecting users privacy~\cite{DBLP:conf/emnlp/WuLXDW0X23}, etc.

Compared with previous studies~\cite{DBLP:conf/nips/MengBAB22,DBLP:conf/iclr/MengSABB23,DBLP:conf/iclr/MitchellLBFM22,DBLP:conf/emnlp/ZhongWMPC23,DBLP:journals/corr/abs-2307-12976,DBLP:journals/corr/abs-2310-10322,DBLP:conf/emnlp/YaoWT0LDC023} that are the most relevant to our work, a main difference should be highlighted. 
These approaches generally focus on determining if the new target
knowledge has been successfully memorized, while this study explores 
the perturbation and impact of editing on neighboring knowledge.
To the best of our knowledge, this paper makes the first attempt to introduce a new metric of \emph{additivity}, build a benchmark for assessing the degree of perturbation to neighboring knowledge
when appending knowledge, and propose a general framework to mitigate the neighboring perturbations.
\section{Preliminary}

\subsection{Querying Factual Knowledge in LLMs}
Following previous works~\cite{DBLP:conf/nips/MengBAB22,DBLP:conf/emnlp/ZhongWMPC23,DBLP:journals/corr/abs-2312-05497,DBLP:conf/emnlp/YaoWT0LDC023,DBLP:journals/corr/abs-2310-10322}, we study factual knowledge of the
form (\(s, r, o)\), consisting of a subject $s$, a relation $r$, and an object $o$ (e.g., $s$ = Eiffel Tower, $r$ = location, $o$ = Paris).
Besides, we also follow them to employ discrete prompts to test whether the knowledge is in a language model.
Specifically, a natural language template $t_r(\cdot)$ is constructed for each relation $r$, which is then combined with a subject $s$ as input to generate a prompt (question or cloze-style statement) $t_r(s)$.
For instance, given the subject ``Eiffel Tower'' and the relation ``location'', we can form a cloze sentence ``The Eiffel Tower is located in''.

\subsection{Knowledge Editing} \label{pre-ke}
The goal of knowledge editing is to incorporate new facts into model parameters without retraining.
Previous works focus on altering the original knowledge stored in LLMs as $(s, r, o)$ $\rightarrow $ $(s, r, o^{*})$, which only cares about the new target knowledge. 
However, the potential impact of the editing operation of appending new knowledge on its neighboring knowledge remains unclear.
Therefore, this paper explores the scenarios where a subject has multiple corresponding objects under a given relation.
In detail, appending knowledge aims at incorporating a new object to the set of original objects, retaining original objects, and not introducing false objects:
$(s, r, \{o_1, o_2,..., o_N\})$ $\rightarrow $ $(s, r, \{o_1, o_2,..., o_N, o^{*}\})$.
In this paper, for a new piece of knowledge to append $(s, r, o^{*})$, its \emph{neighboring knowledge} is defined as $(s, r, O)$ where $O=\{o_1, o_2,..., o_N\}$.
Define an editing fact as $e = (s ,r, O, o^{*})$  and given an unedited model  $\mathcal{F}$ , the edited language model is gained by using a model editor $K$, $\mathcal{F}^{*}=K(\mathcal{F}, e)$. 

To evaluate the performance of editing methods in appending knowledge, the previous three metrics are still utilized.

\textbf{Efficacy} validates whether the edited models could recall the appended fact under the sole editing prompt $p$. 
The assessment is based on Efficacy Score \textbf{(ES)} representing as: $\mathbbm{1}[P_{\mathcal{F}^*}(o^{*} \,|\, p) > min\{P_{\mathcal{F}^*}(o\,|\,p)\, |\, o\in O\}]$\footnote{Previous works care about whether the probability surpasses the original $o$. Here $O$ is a set, so we make a slight adaptation.}.

\textbf{Generalization} verifies whether the edited models could recall the appending fact under the paraphrase prompts $\mathcal{P}^{G}$ via Generalization Score \textbf{(GS)}:
$\mathbb{E}_{p \in \mathcal{P}^{G}}[\mathbbm{1}[\, P_{\mathcal{F}^*}(o^{*} \,|\, p) > min\{P_{\mathcal{F}^*}(o\,|\,p)\, |\, o\in O\}]\,]$.

\textbf{Locality} verifies whether the output of the edited models
for inputs out of editing scope remains unchanged under the locality prompts $\mathcal{P}^{L}$ via Locality Score \textbf{(LS)}:
$\mathbb{E}_{p_l \in \mathcal{P}^{L}}[\mathbbm{1}[\,P_{\mathcal{F}^*}(o_l \,|\, p_l) > P_{\mathcal{F}^*}(o^{*}\,|\, p_l)]\,]$, where $o_l$ was the original answer of $p_l$. 
$\mathbbm{1}$ is the indicator function.

For example, given an editing fact ($s$ = Apple, $r$ = products, $O$ = \{AirPods 3, MacBook Air,..., iPhone 14\}, $o^{*}$=iPhone 15), the editing prompt $p$ could be ``What are the products of Apple?'' and a paraphrase prompt could be ``What items does Apple produce?'' 
A locality prompt and its original answer could be ``Which company developed Windows?'' and ``Microsoft'' respectively.

\section{Definition of Additivity} \label{add}
Given a subject $s$, a relation $r$, a composed question $t_r(s)$, original objects $O$, and an object to append $o^{*}$, it is unclear whether the original correct knowledge $K^c = (s, r, \{o_1, o_2,..., o_N\})$ is still retained,
and part of the false knowledge $K^{f} = (s, r, \{o_{f1}, o_{f2},..., o_{fM}\})$ is unintentionally included after editing, where $\{o_{f1}, o_{f2},..., o_{fM}\} \cap \{o_1, o_2,..., o_N, o^{*}\} = \varnothing$.
Therefore, a new metric termed \emph{additivity} is designed to evaluate the degree of perturbation to neighboring knowledge, which is calculated in terms of \emph{relative ranking} and \emph{absolute probability change} of objects.

\vspace{-3mm}
\paragraph{Relative ranking of objects}
We assume that the minimum probability of correct knowledge $K^c$ should be larger than the maximum probability of false knowledge $K^{f}$ before and after editing.
Otherwise, there is a risk of forgetting original correct knowledge or introducing false knowledge.
Intuitively, the maximum probability of false knowledge $P_{f}^{max} = \max_{i=1}^{M} P_{\mathcal{F}^{*}}(o_{fi} | t_r(s))$ is used as a threshold to calculate the proportion of the list of original correct objects of $K^c$ whose probabilities are below this threshold.
Mathematically, we have \emph{Ranking Forgetting Factor} (RFF):
\begin{equation}
\begin{small}
    \begin{aligned}
        & RFF = \\
        & \frac{ \sum_{i=1}^{N} [\mathbbm{1} \{P_{\mathcal{F}^{*}}(o_{i} | t_r(s))<P_{f}^{max}\} * \sigma(P_{\mathcal{F}^{*}}(o_{i} | t_r(s)))]}{\sum_{i=1}^N \sigma(P_{\mathcal{F}^{*}}(o_{i} | t_r(s)))},
    \end{aligned}
\end{small}
\end{equation}
\vspace{-3mm}

$\sigma$ is the sigmoid function. 
Here, $\sigma(P)$ of each original correct knowledge is regarded as its weight, since knowledge with larger probability should weigh more and the sigmoid function can also be used for smoothing. 

Similarly, the minimum probability of original correct knowledge 
$P_{c}^{min} = \min_{i=1}^{N} P_{\mathcal{F}^{*}}(o_{i} | t_r(s))$ is used as a threshold to calculate the proportion of the list of sampled false objects of $K^f$ whose probabilities are above this threshold.
Then, \emph{Ranking Noise Factor} (RNF) is defined as:
\vspace{-2mm}
\begin{equation}
\begin{small}
    \begin{aligned}
        & RNF = \\ 
        & \frac{ \sum_{i=1}^{M} [\mathbbm{1} \{P_{\mathcal{F}^{*}}(o_{fi} | t_r(s))>P_{c}^{min}\} * \sigma(P_{\mathcal{F}^{*}}(o_{fi} | t_r(s)))]}{\sum_{i=1}^M \sigma(P_{\mathcal{F}^{*}}(o_{fi} | t_r(s)))},
    \end{aligned}
\end{small}
\end{equation}

\vspace{-3mm}
\paragraph{Absolute probability change of objects} 
In addition to satisfying the assumption of relative ranking of objects, it is also necessary to characterize their absolute probability changes.
Even if the relative ranking remains unchanged, substantial harm is inflicted upon the model if the absolute probability changes unexpectedly.
Therefore, \emph{Correct Probability Change} (CPC) is introduced to characterize this issue, which is defined as the ratio of the mean probability of correct knowledge after and before editing:
\begin{equation}
\begin{small}
        \begin{aligned}
        CPC = \frac{\sum_{i=1}^N P_{\mathcal{F}^{*}}(o_{i}| t_r(s))}{\sum_{i=1}^N P_{\mathcal{F}}(o_{i}| t_r(s))}.
        \end{aligned}
\end{small}
\end{equation}
\vspace{-4mm}

Similarly, \emph{False Probability Change} (FPC) is defined as the ratio of the mean probability of false knowledge after and before editing:
\begin{equation}
\begin{small}
        \begin{aligned}
        FPC = \frac{\sum_{i=1}^M P_{\mathcal{F}^{*}}(o_{fi}| t_r(s))}{\sum_{i=1}^M P_{ \mathcal{F}}(o_{fi} | t_r(s))}.
        \end{aligned}
\end{small}
\end{equation}
\vspace{-4mm}

\paragraph{Aggregation} 
Finally, we aim to aggregate these two dimensions of relative ranking and absolute probability change into a unified metric, providing a more comprehensive representation of the detrimental impact induced by appending knowledge.
The \emph{Additive Forgetting Factor} (AFF) is defined as the degree to which the original correct knowledge is forgotten:
\begin{equation}
\begin{small}
        \begin{aligned}
        AFF = 1 - (1 - RFF)*min\{1, CPC\}. \\
        \end{aligned}
\end{small}
\end{equation}
This definition means that if the probability of correct knowledge does not decrease after editing ($CPC>=1$), AFF equals RFF. 
Otherwise, the adverse effects of CPC and RFF would combine, resulting in AFF surpassing RFF.
The AFF spans from 0 to 1, reflecting the extent to which the original correct knowledge is forgotten amidst neighboring perturbations.
A higher AFF value corresponds to a more substantial negative impact.
Similarly, the other aggregated metric \emph{Additive Noising Factor} (ANF) is defined as the degree to which false knowledge is introduced:
\begin{equation}
\begin{small}
        \begin{aligned}
        ANF = 1 - (1 - RNF)*min\{1, \frac{1}{FPC}\}. \\
        \end{aligned}
\end{small}
\end{equation}
The ANF spans from 0 to 1, reflecting the extent to which the false knowledge is introduced in neighboring perturbations. 
A higher ANF value corresponds to a more substantial negative impact.

\section{PEAK: Perturbation Evaluation of Appending Knowledge}
This paper designs the PEAK benchmark to assess the extent of perturbation to neighboring knowledge during editing.
It comprises two datasets of PEAK-CF and PEAK-T.
The former is designed as a counterfact dataset for the evaluation of knowledge editing methods on counterfactual appending following~\citet{DBLP:conf/nips/MengBAB22} and ~\citet{DBLP:journals/corr/abs-2310-10322}.
The latter is based on temporal knowledge edits of changes in the real-world following~\citet{DBLP:conf/emnlp/ZhongWMPC23}.

\subsection{Data Construction of PEAK-CF}

\paragraph{Aggregating facts} This dataset is constructed based on Wikidata~\cite{DBLP:journals/cacm/VrandecicK14}, a knowledge base containing millions of fact triples.
First, we manually chose a total of 32 relations, where a subject has multiple corresponding objects under each relation.
Then we collected all triples for each relation and classified triples of the same subject $s$ and relation $r$ together, denoted as $F_m(r)= \{(s, r, O )\mid O=\{o_1, o_2,..., o_N\})\}$.
For each relation $r$, ChatGPT (gpt-3.5-turbo) was used to generate some templates $T(r)$ expressing the same semantics.
The subject $s$ can be replaced to form the full prompts: $\mathcal{P}(s,r)$ = \{$t_r(s)$ $\mid$ $t_r(\cdot) \in T(r)$\}. 
For example, a template for ($r$ = product) might be “\{\} has products like”, where “Apple” substitutes “\{\}”.

\vspace{-2mm}
\paragraph{Constructing counterfactual edits} Following previous works~\cite{DBLP:conf/nips/MengBAB22,DBLP:conf/emnlp/YaoWT0LDC023,DBLP:journals/corr/abs-2310-10322}, counterfactual edits were built in this dataset.
The previous three metrics are used in this dataset.
Given a set of triples $(s, r, O) \in F_m(r)$ and an object $o^{*}$, where $o^{*} \notin O$ and $(s, r, o^{*})$ is a counterfact,
an edit is represented as  $\mathcal{E}=\left\{s, r, O, o^{*}, p\right\}$ to test \emph{efficacy}, where the editing prompt $p \in \mathcal{P}(s, r)$. 
Besides, to test \emph{generalization}, a set of two semantically-equivalent paraphrase prompts $\mathcal{P}^{G}$ are sampled from $\mathcal{P}(s, r) \backslash\left\{p\right\}$.
Moreover, to test \emph{locality}, a set of triples outside the scope of editing is selected: $\mathcal{S}=\left\{\left(s^{\prime}, r^{\prime}, o^{\prime}\right)\right\}$. 
For example, select  ($s$= Olympic Games,  $r$=  host city,  $o$=  Paris),  $\mathcal{S}$  might contain triple like ($s^{\prime}$= France,  $r^{\prime}$=  shares border,  $o^{\prime}$= Germany).
Then a set of prompts $\left\{\mathcal{P}\left(s^{\prime}, r^{\prime}\right) \mid (s^{\prime},r^{\prime},o^{\prime}) \in \mathcal{S}\right\}$ is constructed to sample the locality prompts $\mathcal{P}^{L}$.

\vspace{-2mm}
\paragraph{Sampling false answers} For the \emph{additivity} proposed in this paper, 
given a factual question $t_r(s)$ (e.g., the editing prompt $p$),
both the original correct answers and sampled false answers are utilized.
$O$ is the list of original correct answers, while the false answers were sampled in two settings \emph{Hard} and \emph{Random} (sampling details are in Appendix~\ref{append-cons}).
For the \emph{Hard} setting, some objects that establish direct relations with the new object $o^*$ are selected, represented as $O_h=\{o_{h1}, o_{h2},..., o_{hM}\}$, 
where $O_h \cap O=\varnothing$.
Similarly, for the \emph{Random} setting, some objects $O_r =\{o_{r1}, o_{r2},..., o_{rM}\}$ that are semantically distant from the new object are sampled.
Intuitively, false answers in the \emph{Hard} setting are more easily introduced into the model compared to answers in the \emph{Random} setting.

\vspace{-2mm}
\paragraph{Filtering correct and false answers} Subsequently, according to additivity defined in Section~\ref{add}, for $t_r(s)$, the minimum probability of correct answers $O$ should be larger than the maximum probability of false answers $O_{h}$ and $O_{r}$ before and after editing.
Therefore, before editing, we filtered out correct answers with probabilities below a certain threshold in the original model.
Then false answers with probabilities greater than the minimum probability of the selected correct answers were also filtered out.

\begin{table}[t]
\begin{footnotesize}
\renewcommand\arraystretch{1}
\caption{An example in the PEAK-T dataset. $a$ and $a^*$ in different rows represent the original answer and desired answer after editing respectively. $\cup$ refers to the set union operation.} \label{example}
\centering
\setlength{\tabcolsep}{0.9mm}{
\begin{tabular}{ll}
\toprule  
\multirow{5}{*}{$\mathcal{E}$} & $(s,r)$: (Olympic Winter Games, host country) \\
& $O$: \{France, United States, ..., South Korea\}  \\
& $o^*$:  China \\ 
& $p$: What are the host countries of Olympic Winter Games?\\
& ($a$: $O$, $a^{*}$: $O \cup \{o^*$\}) \\
\midrule
\multirow{3}{*}{$\mathcal{P}^{G}$} & The host countries of Olympic Winter Games are  \\
& Which nations have hosted the Winter Olympics? \\ 
& ($a$: $O$, $a^{*}$: $O \cup \{o^*$\}) \\
\midrule
\multirow{2}{*}{$\mathcal{P}^{L}$} & Big Ben is located in ($a$ \& $a^*$: London)\\
& The headquarters of Apple is in ($a$ \& $a^*$: California)\\ 
\midrule
\multirow{1}{*}{$O_h$} & \{Japan, Shanghai, Hong Kong, ..., Russia\} \\
\midrule
\multirow{1}{*}{$O_r$} & \{Uruguay, Mozambique, Fiji, ..., Tripoli\}  \\
\bottomrule 
\end{tabular}}
\vspace{-3mm}
\end{footnotesize}
\end{table}

\subsection{Data Construction of PEAK-T}
This dataset focuses on temporal-based, real-world edits, where each edit is factually correct and occurs after the model is trained.
Following~\citet{DBLP:journals/corr/abs-2312-05497}, it is built upon
YAGO~\cite{DBLP:conf/cidr/MahdisoltaniBS15}, 
a knowledge base containing fact triples extracted from Wikipedia, enriched with WordNet, GeoNames, and other data sources, and contains the time when the facts occurred.
We sampled facts in YAGO that occurred after 2021, which took place following the release of GPT-2 XL and GPT-J.
Due to the limited number of relation types, 9 relations have been selected.
Then the construction is similar to PEAK-CF.

\subsection{Dataset Summary}

\paragraph{Dataset Format}
As shown in Table~\ref{example}, each instance in the PEAK benchmark is represented as a tuple ($\mathcal{E}$, $\mathcal{P}^{G}$, $\mathcal{P}^{L}$, $O_h$, $O_r$),
where $\mathcal{E}$ comprises a piece of new knowledge and original correct knowledge.
$\mathcal{P}^{G}$ and $\mathcal{P}^{L}$ are the prompts utilized to validate generalization and locality respectively.
$O_h$ and $O_r$ represent the sampled false answer lists for the factual questions $p$ and $\mathcal{P}^{G}$ under the \emph{Hard} and \emph{Random} settings respectively. 
$a$ and $a^{*}$ denote the original answer and desired
answer after editing respectively.

\begin{table}[t]
\small
\renewcommand\arraystretch{1.1}
\caption{The statistics of PEAK-CF and PEAK-T datasets.} \label{stat}
\centering
\setlength{\tabcolsep}{1.8mm}{
\begin{tabular}{llcc}
\toprule
& Type                 & PEAK-CF & PEAK-T \\
\midrule
& Editing examples     & 1,962   & 993    \\
& Correct answers      & 26,512  & 7,759  \\
& Paraphrase prompts   & 3,344   & 1,995  \\
& Neighborhood prompts & 10,760  & 6,207  \\
& False answers (hard)  & 24,043  & 7,879  \\
& False answers (random)  & 19,616  & 9,929  \\
\bottomrule 
\end{tabular}}
\vspace{-3mm}
\end{table}

\vspace{-3mm}
\paragraph{Dataset Statistics}
Table~\ref{stat} summarizes the statistics of the PEAK-CF and PEAK-T datasets.
To verify generalization and locality, there is at least one prompt for each instance. 
Besides, each fact question has about an average of 10 correct and 20 false answers.
Generally speaking, these two datasets contain counterfactual edits and temporal edits respectively, and are used to study neighborhood perturbations of edited models.
Readers can refer to Appendix~\ref{append-temple} for the details of relations and prompts.

\begin{table*}[ht]
\footnotesize
\centering
\caption{Evaluation results (\%) of the PEAK-CF dataset.
``hard'' refers to the additivity calculated using false answers in the \emph{Hard} setting, while ``ran'' refers to that in the \emph{Random} setting.
$\uparrow$ indicates that a higher value corresponds to better performance, while $\downarrow$ is the opposite. 
Numbers marked with $\dagger$ indicate statistically significant improvements in the method coupled with APP over the original method (t-test with \(p\)-value \textless \(0.05\)).
Since KN has no loss to optimize, APP is not applied to it.
Due to space limitation, the results of LLaMA-2 (13B) were put in Appendix~\ref{append-13b}.
}
\label{cfresults}
\renewcommand\arraystretch{1.05}
\setlength{\tabcolsep}{1mm}{
\resizebox{0.98\linewidth}{!}{
\begin{tabular}{lccccccc|ccccccc}
\toprule
\multirow{3}{*}{\textbf{ Editor }} 
& \multicolumn{7}{c}{\textbf { GPT-2 XL (1.5B) }} 
& \multicolumn{7}{c}{\textbf { LLaMA-2 (7B)  }} 

 \\
\cmidrule(r){2-8}
\cmidrule(r){9-15}
&\multicolumn{3}{c}{\textbf { Previous }} 
&\multicolumn{2}{c}{\textbf { Additivity (hard) }} 
&\multicolumn{2}{c}{\textbf { Additivity (ran) }} 
&\multicolumn{3}{c}{\textbf { Previous }} 
&\multicolumn{2}{c}{\textbf { Additivity (hard) }} 
&\multicolumn{2}{c}{\textbf { Additivity (ran) }} 
\\
\cmidrule(r){2-4}
\cmidrule(r){5-6}
\cmidrule(r){7-8}
\cmidrule(r){9-11}
\cmidrule(r){12-13}
\cmidrule(r){14-15}
& $\mathrm{ES} \uparrow$ & $\mathrm{GS} \uparrow$  & $\mathrm{LS} \uparrow$ & $\mathrm{AFF} \downarrow$ & $\mathrm{ANF} \downarrow$ & $\mathrm{AFF} \downarrow$ & $\mathrm{ANF} \downarrow$  
& $\mathrm{ES} \uparrow$ & $\mathrm{GS} \uparrow$  & $\mathrm{LS} \uparrow$ & $\mathrm{AFF} \downarrow$ & $\mathrm{ANF} \downarrow$ & $\mathrm{AFF} \downarrow$ & $\mathrm{ANF} \downarrow$ \\
 \midrule[0.5pt]
 \text {FT} &90.99	&64.68	&87.71	&78.86	&59.04	&69.29	&37.12 &98.27	&87.72	&68.48	&74.51	&60.62	&64.22	&38.72

 \\
  \text {KN} &26.48	&26.08	&89.08	&44.71	&29.47	&41.23	&22.96  &46.67	&46.45	&66.16	&45.63	&40.51	&43.72	&35.64

  \\
  \text {MEND} &97.46	&84.87	&42.46	&30.15	&33.97	&12.86	&11.62  &95.22	&86.02	&47.47	&35.39	&33.51	&21.16	&14.61

  \\
  \text {MEMIT} &72.90	&63.08	&98.44	&51.49	&52.11	&31.71	&17.22  &99.95	&\textbf{99.11}	&92.12	&94.26	&84.14	&86.03	&53.61

  \\
  \text {ROME} &\textbf{99.27}	&\textbf{94.05}	&90.57	&87.44	&69.80	&75.16	&35.58  &99.69 &	97.74	&94.40	&93.05	&82.47	&83.80	&52.25
  \\
\hline\\[-2mm]\hline
\rule{0pt}{9pt} 

 \text {FT+APP}  &87.89	&59.30	&89.21	&70.48$^{\dagger}$	&49.73$^{\dagger}$ &59.85$^{\dagger}$	&29.76$^{\dagger}$
&98.01	&85.24	&73.55$^{\dagger}$	&68.44$^{\dagger}$	&49.62$^{\dagger}$	&57.93$^{\dagger}$	&33.34$^{\dagger}$

 \\
  \text {MEND+APP} &94.25	&81.02	&45.51$^{\dagger}$	&\textbf{26.27}$^{\dagger}$	&\textbf{28.81}$^{\dagger}$	&\textbf{12.21}	&\textbf{11.07}
&92.86	&82.46	&51.43$^{\dagger}$	&\textbf{32.55}$^{\dagger}$	&29.73$^{\dagger}$	&\textbf{20.44}	&13.99

  \\
  \text {MEMIT+APP} &69.75	&59.26	&\textbf{98.66}	&39.89$^{\dagger}$	&43.41$^{\dagger}$	&22.14$^{\dagger}$	&13.58$^{\dagger}$
&99.42	&96.51	&94.81$^{\dagger}$	&38.83$^{\dagger}$	&\textbf{17.41}$^{\dagger}$	&32.65$^{\dagger}$	&\textbf{11.69}$^{\dagger}$
  \\
  \text {ROME+APP} &96.15	&87.17	&93.27$^{\dagger}$	&40.45$^{\dagger}$	&31.60$^{\dagger}$	&25.92$^{\dagger}$	&12.61$^{\dagger}$
&\textbf{100.00}	&94.24	&\textbf{96.22}$^{\dagger}$	&43.11$^{\dagger}$	&20.14$^{\dagger}$	&36.86$^{\dagger}$	&13.74$^{\dagger}$
  \\
\bottomrule
\end{tabular}}}
\vspace{-3mm}
\end{table*}

\section{Appending via Preservation and Prevention}
The neighboring perturbations in the process of appending knowledge may lead to the forgetting of original correct knowledge, as well as the unintentional inclusion of noise.
A plug-and-play framework APP (\textbf{A}ppending via \textbf{P}reservation and \textbf{P}revention) is proposed to improve existing editing methods to mitigate this detriment in editing.

Given the new knowledge to append $(s, r, o^*)$ and an editing prompt $p$, current editing methods usually define an editing objective $\mathcal{L}_{e}(o^*, \theta_\mathcal{F})$ to incorporate the new knowledge, where $\theta_\mathcal{F}$ denotes the parameters to be updated in the unedited model $\mathcal{F}$.
APP designs a set of editing objectives which can be coupled with $\mathcal{L}_{e}(o^*, \theta_\mathcal{F})$, to minimize the probability perturbations of both neighboring correct and incorrect knowledge. 
On the one hand, the editing objective $\mathcal{L}_{1}(O, O_{h},\theta_\mathcal{F})$ is designed to maintain a certain margin between the probabilities of original correct answers $O$ and those of false answers $O_{h}$ for the question $t_r(s)$ via the Hinge Loss~\cite{DBLP:conf/nips/GentileW98,wu2007robust} as follows:\footnote{$O_{h}$ is adopted due to hard negative sampling.}
\vspace{-2mm}
\begin{equation}
    \begin{aligned}
        \mathcal{L}_{1}(O, O_{h}, \theta_\mathcal{F}) &= \frac{1}{NM}\sum_{i=1}^{N}\sum_{j=1}^{M} max\{0, \: \mathcal{M}\\
         &-\log P_{\mathcal{F}^{\prime}}(o_{i} \,|\, p)  + \log P_{\mathcal{F}^{\prime}} (o_{hj} \,|\, p)\},
    \end{aligned}
\end{equation}
where $\mathcal{F}^{\prime}$ is the intermediate model of the editing process.
$N$ and $M$ represent the number of elements in $O$ and $O_{h}$ respectively.
This equation implies that the log probabilities of correct answers are encouraged to be larger than those of false answers by at least a certain margin $\mathcal{M}$.

On the other hand, it involves ensuring that the absolute probabilities of correct answers do not decrease
while those of false answers do not increase during editing, which can be conceptualized as two objectives $\mathcal{L}_{2}(O,\theta_\mathcal{F})$ and $\mathcal{L}_{3}(O_h,\theta_\mathcal{F})$ respectively as:
\begin{equation}
    \begin{aligned}
        \mathcal{L}_{2}(O,\theta_\mathcal{F}) = & \frac{1}{N} \sum_{i=1}^{N} max\{0, \,  \\
        & \log P_{\mathcal{F}}(o_{i} \,|\, p) - \log P_{\mathcal{F}^{\prime}}(o_{i} \,|\, p)\},
    \end{aligned}
\end{equation}

\vspace{-6mm}

\begin{equation}
    \begin{aligned}
        \mathcal{L}_{3}(O_h,\theta_\mathcal{F}) = & \frac{1}{M} \sum_{i=1}^{M}max\{0, \,  \\
        & \log P_{\mathcal{F}^{\prime}}(o_{hi} \,|\, p)- \log P_{\mathcal{F}}(o_{hi} \,|\, p)\}, \\
    \end{aligned}
\end{equation}

$\mathcal{L}_{2}(O,\theta_\mathcal{F})$ means that if the probability of the correct answer decreases during editing, the loss is equal to the log probability decrease value, otherwise it is 0.
$\mathcal{L}_{3}(O_h,\theta_\mathcal{F})$ is designed similarly. 
Finally, these proposed objectives are jointly optimized with the original editing objective in each method $\mathcal{L}_{e}(o^*, \theta_\mathcal{F})$ as:

\vspace{-1mm}
\begin{equation}
    \begin{aligned}
        \mathcal{L} =& \mathop{\min}\limits_{\theta_\mathcal{F}} \mathcal{L}_{e}(o^*,\theta_\mathcal{F}) + \alpha \mathcal{L}_{1}(O, O_{h},\theta_\mathcal{F}) \\ 
                    &+ \beta \mathcal{L}_{2}(O,\theta_\mathcal{F}) +\gamma \mathcal{L}_{3}(O_{h},\theta_\mathcal{F}).
        \label{final}
    \end{aligned}
\vspace{-1mm}
\end{equation}

Here $\alpha,\beta,\gamma$ $>=0$ are hyperparameters.
Appendix~\ref{exampleapp} provides an example illustrating how the proposed APP method is coupled with a selected editing method for clarity.

\section{Experiments} \label{experiment_baseline}

In this section, experiments were conducted to demonstrate the neighborhood perturbations of existing editing methods based on four representative LLMs.
Furthermore, the proposed APP method was coupled with these editing methods to mitigate the neighborhood perturbations.
Finally, comprehensive analyses were carried out to further verify the effectiveness of the proposed framework.

\subsection{Base LLMs and Editing Methods} \label{baselines}
Four popular LLMs were used for experiments.
Considering limited computing resources, the PEAK-CF dataset was conducted on \textbf{GPT-2 XL} (1.5B)~\cite{radford2019language}, \textbf{LLaMA-2} (7B) and \textbf{LLaMA-2} (13B)~\cite{DBLP:journals/corr/abs-2307-09288}.
Since there is little data in YAGO after LLaMA-2 release date (2023-7), The PEAK-T was conducted on \textbf{GPT-2 XL} and \textbf{GPT-J} (6B)~\cite{wang2021gpt}.
Five popular knowledge editing methods were selected as the baselines including
\textbf{FT}~\cite{DBLP:journals/corr/abs-2012-00363},
\textbf{KN}~\cite{DBLP:conf/acl/DaiDHSCW22},
\textbf{MEND}~\cite{DBLP:conf/iclr/MitchellLBFM22},
\textbf{ROME}~\cite{DBLP:conf/nips/MengBAB22}, and \textbf{MEMIT}~\cite{DBLP:conf/iclr/MengSABB23}.
Readers can refer to Appendix~\ref{append_baselines}  for the details of these editing methods.

\begin{table*}[ht]
\footnotesize
\centering
\setlength{\abovecaptionskip}{0.1cm} 
\caption{Evaluation results (\%) of the PEAK-T dataset.}
\label{tresults}
\renewcommand\arraystretch{1.05}
\setlength{\tabcolsep}{1.2mm}{
\resizebox{0.98\linewidth}{!}{
\begin{tabular}{lccccccc|ccccccc}
\toprule
\multirow{3}{*}{\textbf{ Editor }} 
& \multicolumn{7}{c}{\textbf { GPT-2 XL (1.5B) }} 
& \multicolumn{7}{c}{\textbf { GPT-J (6B)  }} 

 \\
\cmidrule(r){2-8}
\cmidrule(r){9-15}
&\multicolumn{3}{c}{\textbf { Previous }} 
&\multicolumn{2}{c}{\textbf { Additivity (hard) }} 
&\multicolumn{2}{c}{\textbf { Additivity (ran) }} 
&\multicolumn{3}{c}{\textbf { Previous }} 
&\multicolumn{2}{c}{\textbf { Additivity (hard) }} 
&\multicolumn{2}{c}{\textbf { Additivity (ran) }} 
\\
\cmidrule(r){2-4}
\cmidrule(r){5-6}
\cmidrule(r){7-8}
\cmidrule(r){9-11}
\cmidrule(r){12-13}
\cmidrule(r){14-15}
& $\mathrm{ES} \uparrow$ & $\mathrm{GS} \uparrow$  & $\mathrm{LS} \uparrow$ & $\mathrm{AFF} \downarrow$ & $\mathrm{ANF} \downarrow$ & $\mathrm{AFF} \downarrow$ & $\mathrm{ANF} \downarrow$  
& $\mathrm{ES} \uparrow$ & $\mathrm{GS} \uparrow$  & $\mathrm{LS} \uparrow$ & $\mathrm{AFF} \downarrow$ & $\mathrm{ANF} \downarrow$ & $\mathrm{AFF} \downarrow$ & $\mathrm{ANF} \downarrow$ \\
 \midrule[0.5pt]
 \text {FT} &83.31	&59.89	&61.67	&80.44	&38.10	&72.24	&18.76
&95.21	&80.68	&42.74	&87.25	&40.14	&77.09	&14.97

 \\
  \text {KN} &28.39	&27.38	&83.81	&50.85	&24.41	&47.29	&18.73
&45.49	&43.11	&67.05	&93.84	&46.47	&93.29	&40.23
  \\
  \text {MEND} &99.00	&89.90	&8.78	&24.02	&19.78	&12.91	&8.18
&99.72	&87.55 	&10.05 	&24.73	&16.82	&13.66	&5.12

  \\
  \text {MEMIT} &57.92  &46.46	&97.71	&25.92	&31.45	&7.31	&4.19
&100.00	&85.42	&94.00	&50.18	&36.35	&22.85	&2.95

  \\
  \text {ROME} &\textbf{100.00}	&86.77	&92.95	&57.65	&33.84	&36.05	&5.35
&100.00	&\textbf{92.57}	&89.15	&60.48	&38.66	&33.87	&4.02
  \\
\hline\\[-2.0mm]\hline
\rule{0pt}{9pt} 
 \text {FT+APP} &79.03	&53.19	&67.29$^{\dagger}$	&75.24$^{\dagger}$	&31.64$^{\dagger}$	&67.47$^{\dagger}$	&17.75$^{\dagger}$
&90.85	&74.50	&49.53$^{\dagger}$	&82.75$^{\dagger}$	&31.06$^{\dagger}$	&75.02$^{\dagger}$	&13.41$^{\dagger}$

 \\
  \text {MEND+APP} &99.29	&\textbf{90.68}	&8.37	&26.27	&28.81	&12.21	&11.07
&99.72	&87.49	&9.93	&26.38	&17.05	&14.85	&5.12

  \\
  \text {MEMIT+APP} &59.06	&47.68	&\textbf{97.92}	&\textbf{23.30}$^{\dagger}$	&29.00$^{\dagger}$	&\textbf{6.86}	&4.20
&100.00	&83.23	&\textbf{94.36}	&\textbf{21.40}$^{\dagger}$	&\textbf{16.78}$^{\dagger}$	&\textbf{7.54}$^{\dagger}$	&\textbf{1.93}$^{\dagger}$
  \\
  \text {ROME+APP} &99.86	&86.70	&93.65	&28.81$^{\dagger}$	&\textbf{14.84$^{\dagger}$}	&16.79$^{\dagger}$	&\textbf{3.16}$^{\dagger}$
&\textbf{100.00}	&90.83	&89.90	&31.56$^{\dagger}$	&17.15$^{\dagger}$	&16.17$^{\dagger}$	&2.17$^{\dagger}$
  \\

\bottomrule
\end{tabular}}}
\vspace{-3mm}
\end{table*}

\subsection{Evaluation Metrics} \label{metric-define}

Three basic metrics of \emph{Efficacy} (ES), \emph{Generalization} (GS) and \emph{Locality} (LS)~\cite{DBLP:conf/nips/MengBAB22,DBLP:conf/iclr/MengSABB23} defined in Section~\ref{pre-ke} were still adopted to measure each editing method. 
Besides, the proposed \emph{Additivity} was to evaluate the degree of perturbation to neighboring knowledge when appending knowledge.
\textbf{AFF} and \textbf{ANF} designed in Section~\ref{add} were employed to quantify the extent of forgetting original correct answers and the inclusion of noise and incorrect answers respectively.
When assessing the additivity, both the editing and paraphrase prompts $\{p\} \cup \mathcal{P}^{G}$ were utilized without loss of generality.
For all the above metrics, the average results across all edits in each dataset were reported.

\subsection{Results of Existing Editing Methods}
The top five rows of Table~\ref{cfresults} and Table~\ref{tresults} reported the knowledge editing results of current editing methods on PEAK-CF and PEAK-T respectively.
These results were analyzed from the following perspectives.

\textbf{The performance of editing the new target knowledge.} 
Except for KN, editing methods performed well in efficacy (ES) and generalization (GS), showing that most of existing editing methods are capable of effectively appending new target knowledge.
For locality, locate-then-edit methods (KN, ROME, MEMIT) significantly outperformed other methods, demonstrating that they have little interference with irrelevant knowledge.
Moreover, as the model size increased, the performance of a particular editing method exhibited continuous improvement in appending new facts.

\textbf{The perturbations of editing on neighboring knowledge.} 
We were surprised to observe that existing editing methods significantly perturbed the knowledge neighboring to the target knowledge in LLMs after editing, compromising the integrity of original correct knowledge and introducing unintentional noise.
Taking LLaMA-2 edited by ROME on PEAK-CF as an example, despite its superior performance evaluated by previous metrics, it achieved remarkably poor performance of 93.05\% AFF and 82.47\% ANF respectively in the \emph{Hard} negative setting. 
Even in the \emph{Random} setting, they still show poor performance, indicating severe damage to neighboring knowledge.
Meanwhile, edited models generally showed worse performance under the \emph{Hard} setting than those under the \emph{Random} setting in terms of AFF and ANF, demonstrating that the closer the semantic relationship between the false answer and the newly appended answer, the more susceptible it is to perturbation. 
Remarkably, while MEND performed the best in terms of additivity, its relatively low LS indicates a significant disruption to irrelevant knowledge.
Our findings suggest that current editing methods primarily care about the editing performance of new target knowledge, while neglecting the negative perturbations on its neighboring knowledge.

\textbf{Comparison between datasets.} Comparing the results of each method on the two datasets, two conclusions were drawn.
(1) The ES and GS of each method on PEAK-T were lower than those on PEAK-CF, showing that the PEAK-T is more challenging for existing methods to append knowledge without neighboring perturbations.
This could be attributed to the fact that, for PEAK-T, new knowledge originates from the external world and has never been seen by the model. In contrast, PEAK-CF already contains some weak counterfactual associations, making it easier for the model to integrate counterfacts.
(2) Both AFF and ANF on PEAK-T were lower than those on PEAK-CF, suggesting that PEAK-T suffers fewer neighboring perturbations during editing than PEAK-CF.
This is possibly due to the weaker correlation between the sampled false answers and the new external knowledge compared to the original counterfactual knowledge in the model.
Consequently, the perturbations when appending new knowledge in PEAK-T are smaller.

\subsection{Results of APP} \label{appresults}
As shown in the bottom four rows in Table~\ref{cfresults} and Table~\ref{tresults}, \MODELNAME{} was coupled with four editing methods. 
The false answers utilized in APP were from the \emph{Hard} setting.
Detailed hyperparameters of each method can be referred to in Appendix~\ref{append-hyper}.
In general, \MODELNAME{} almost maintained the performance
of appending new knowledge in terms of previous editing metrics, and greatly mitigated the neighborhood perturbations in terms of AFF and ANF under both the \emph{Hard} and \emph{Random} settings.
Particularly, ROME+APP and MEMIT+APP still performed well in appending new knowledge, with the most substantial reduction in perturbations compared to the original editing methods.
Besides, we also found that \MODELNAME{} has also improved the performance of LS, preserving irrelevant knowledge.
These results help conclude that \MODELNAME{} effectively 
preserves the integrity of original correct knowledge and prevents
the inclusion of false knowledge while appending new knowledge.
Despite the notable improvement in additivity, it remains considerably below a satisfactory level, highlighting the severity and complexity of the proposed neighboring perturbations. 
Addressing this challenge will necessitate collaborative efforts from the community.

\subsection{Probability Change of Answers}

\begin{figure}[t]
\centering
\includegraphics[width=0.475\textwidth]{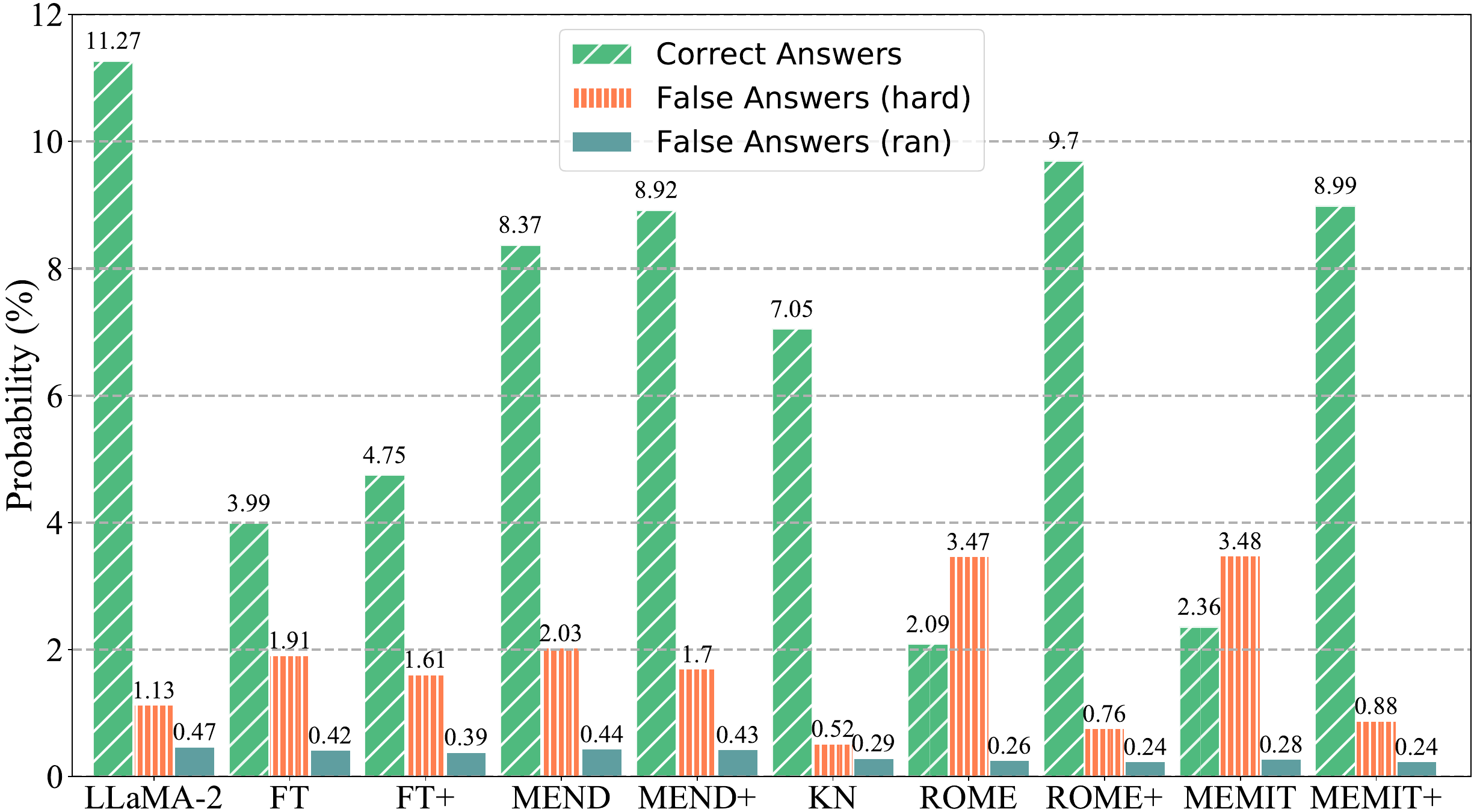}
\vspace{-2mm}
\caption{Average probability of the correct answers $O$ and the false answers $O_h$ (\emph{Hard}) and $O_r$ (\emph{Random}) of LLaMA-2 after editing with different editing methods.
LLaMA-2 refers to the unedited model.
``+'' means this method was coupled with APP. 
} \label{figprob}
\vspace{-4mm}
\end{figure}

To further analyze the neighboring perturbations in edited models and the mechanism of our proposed APP, the average probability of both correct and false answers in LLaMA-2 models edited by different methods on PEAK-CF, is illustrated in Figure~\ref{figprob}.
Two conclusions can be drawn here.

\textbf{Existing editing methods severely perturb probabilities.} 
Compared with the original LLaMA-2, the probabilities of correct answers dropped significantly,
while the probabilities of false answers (\emph{Hard}) increased a lot, especially for MEMIT and ROME.
MEND has shown minimal disturbance, aligning with its superior additivity performance.
Compared to the false answers (\emph{Hard}), the probabilities of false answers (\emph{Random}) undergo a notably smaller shift after editing, indicating that the false answers more closely tied to the new appended answers warrant greater attention.

\textbf{APP effectively mitigates the probability perturbations.} 
After coupling existing editing methods with the proposed APP, the probability perturbations for both correct and false answers were significantly mitigated.
These results vividly explain why APP can effectively mitigate perturbations, thereby preserving the correct knowledge and preventing the inclusion of false knowledge.

\subsection{Ablation study} \label{sec-ablation}
\begin{figure}[t]
  \centering
  \hspace{-1mm}
  \subfigure{
  \includegraphics[width=4.05cm]{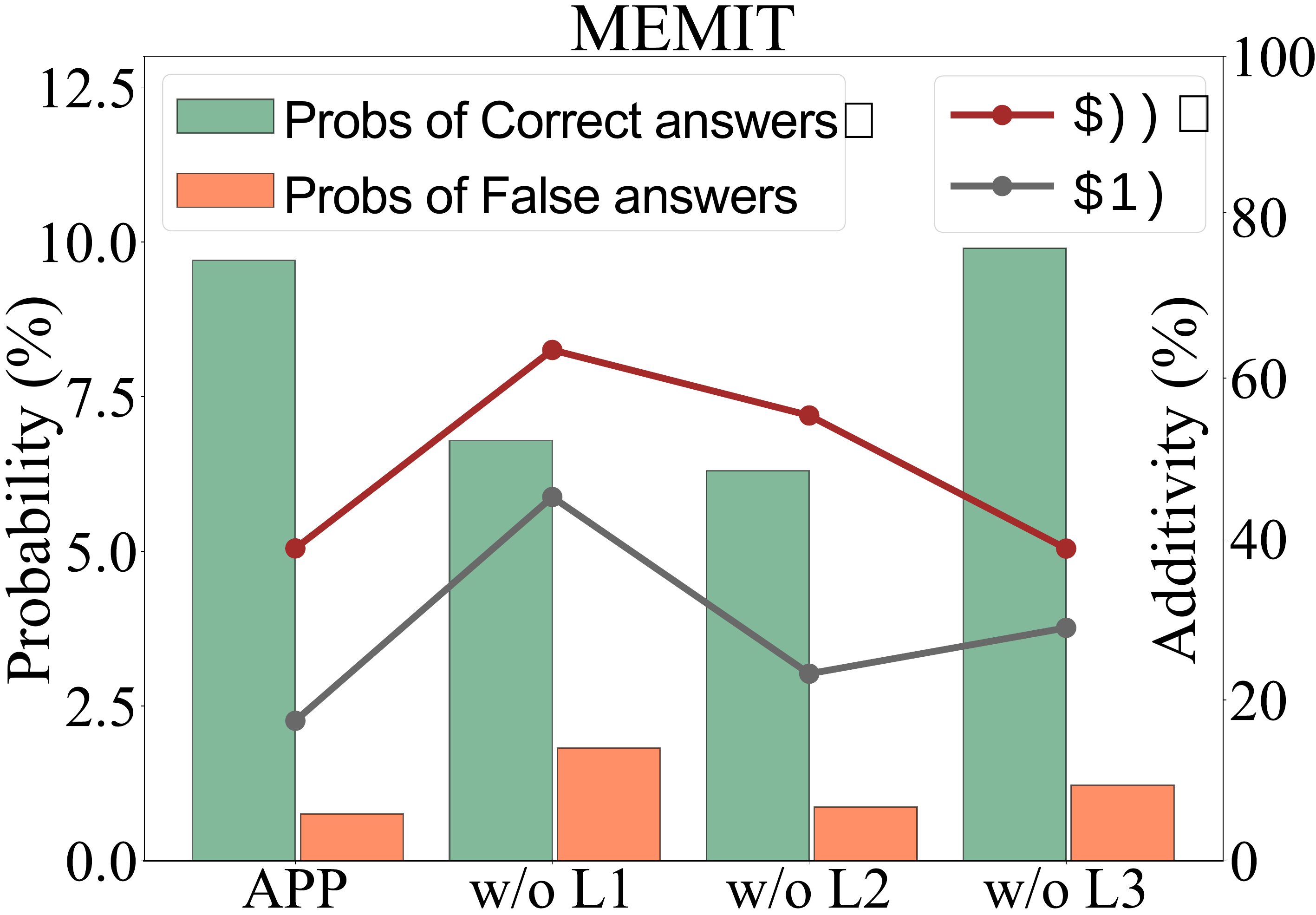}}
  \subfigure{
  \includegraphics[width=3.95cm]{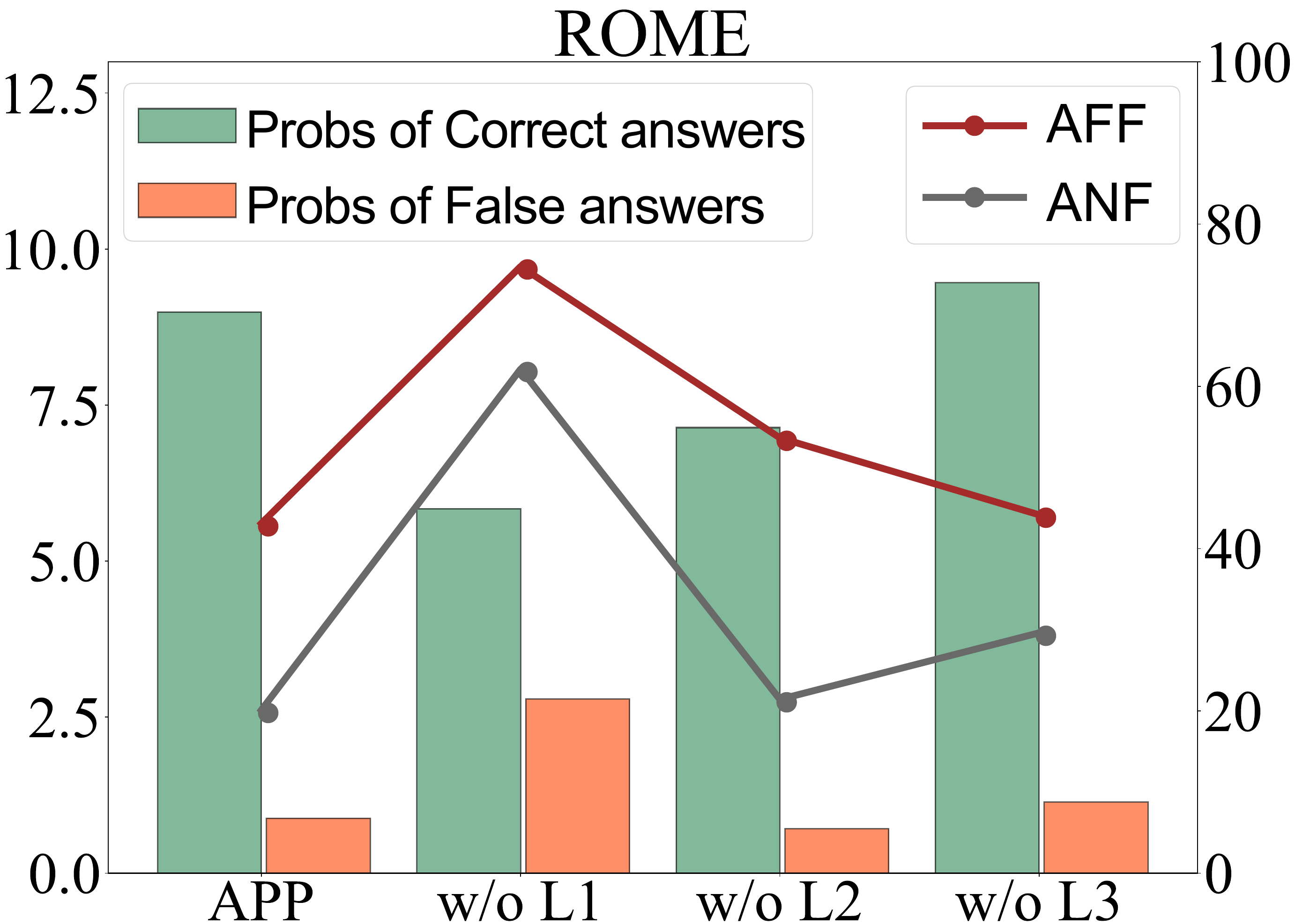}}\hspace{-0mm}
  \vspace{-6mm}
  \caption{
  Ablation analysis of probability and additivity for APP.
  Results were conducted with LLaMA-2 on PEAK-CF dataset.
  Due to page limit, results for other methods are put in Appendix~\ref{append-ablation}.}
  \vspace{-4mm}
  \label{abaltion}
\end{figure}

Ablation tests of removing each editing objective $\mathcal{L}_{1}$, $\mathcal{L}_{2}$ or $\mathcal{L}_{3}$ in APP were conducted to validate their effectiveness.
The AFF and ANF of additivity, as well as the probabilities of correct and false (\emph{Hard}) answers were illustrated in Figure~\ref{abaltion}.
We have two findings.
First, removing any editing objective of the APP led to performance degradation in terms of additivity and probability perturbations, showing their effectiveness.
Second, removing $\mathcal{L}_{1}$ resulted in the most significant performance degradation.
It demonstrated the most importance of $\mathcal{L}_{1}$ in APP, which is designed to maintain a certain margin between the probabilities of correct answers and those of false answers.

\subsection{Effect of Access to All Correct Answers}

The proposed method has access to all the correct answers $O$ during editing, whereas the baseline methods can only have access to the new correct answer $o^*$.
To explore the effect of access to all the correct answers, we let these baseline methods also directly use all correct answers as editing targets when editing.
Specifically, all baseline methods have been extended to utilize the entire set $O$ when editing LLMs to ensure a fair comparison.
The results are shown in Table~\ref{fairer}.
Comparing these results with those of the proposed APP in Table~\ref{cfresults}, we observe that, Despite similar performance in appending new knowledge, the additivity of the new baselines is still significantly higher than that of the proposed APP method, suggesting that APP performs better at maintaining the integrity of the answer list.

\begin{table}[ht]
\footnotesize
\vspace{-3mm}
\centering
\caption{Evaluation results (\%) of fairer baselines on the PEAK-CF dataset with LLaMA-2 (7B). ``+'' refers to the baseline coupled with all correct answers $O$. 
Here only false answers in the \emph{Hard} setting were used.
}
\label{fairer}
\setlength{\tabcolsep}{1.7mm}{
\begin{tabular}{lccccccc|ccccccc}
\toprule
\multirow{3}{*}{\textbf{ Editor }} 
& \multicolumn{5}{c}{\textbf { LLaMA-2 (7B) }} 

 \\
\cmidrule(r){2-6}
&\multicolumn{3}{c}{\textbf { Previous }} 
&\multicolumn{2}{c}{\textbf { Additivity (hard) }} 
\\
\cmidrule(r){2-4}
\cmidrule(r){5-6}
& $\mathrm{ES} \uparrow$ & $\mathrm{GS} \uparrow$  & $\mathrm{LS} \uparrow$ & $\mathrm{AFF} \downarrow$ & $\mathrm{ANF} \downarrow$ \\
 \midrule[0.5pt]

 \text {FT+}  &97.54	&85.93	&71.68	&74.35	&61.83

 \\
  \text {MEND+} &93.26	&79.90	&50.91	&33.92	&32.03

  \\
  \text {MEMIT+} &98.29	&93.07	&94.20	&48.10	&66.72
  \\
  \text {ROME+} &100.00	&93.13	&95.48	&50.27	&64.91
  \\
\bottomrule
\end{tabular}}
\vspace{-3mm}
\end{table}

\subsection{Effect of Number of Neighboring Answers}
\begin{figure}[t]
  \centering
  \subfigure{
  \includegraphics[width=7.09cm]{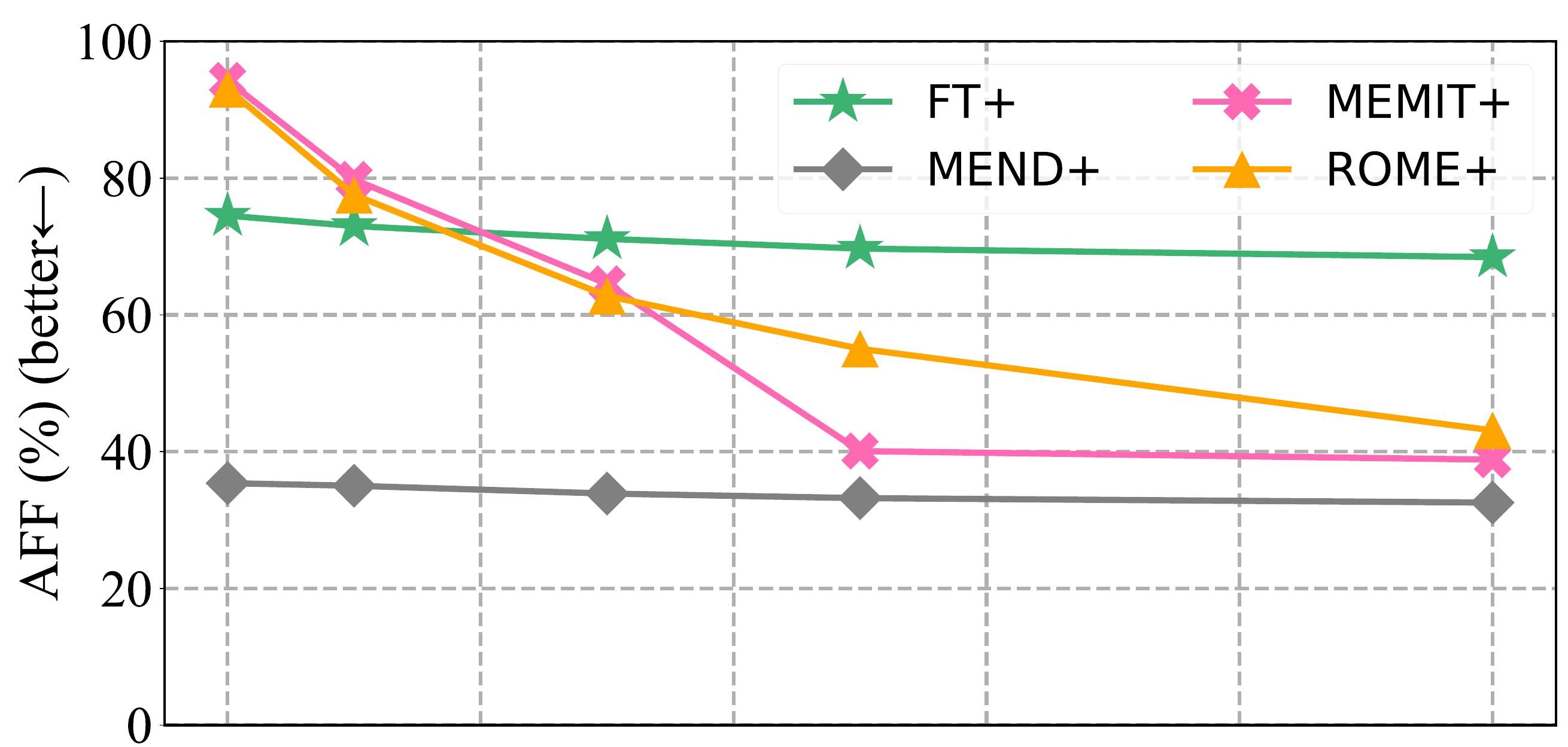}}\hspace{1mm} \vskip -5pt
  \subfigure{
  \includegraphics[width=7cm]{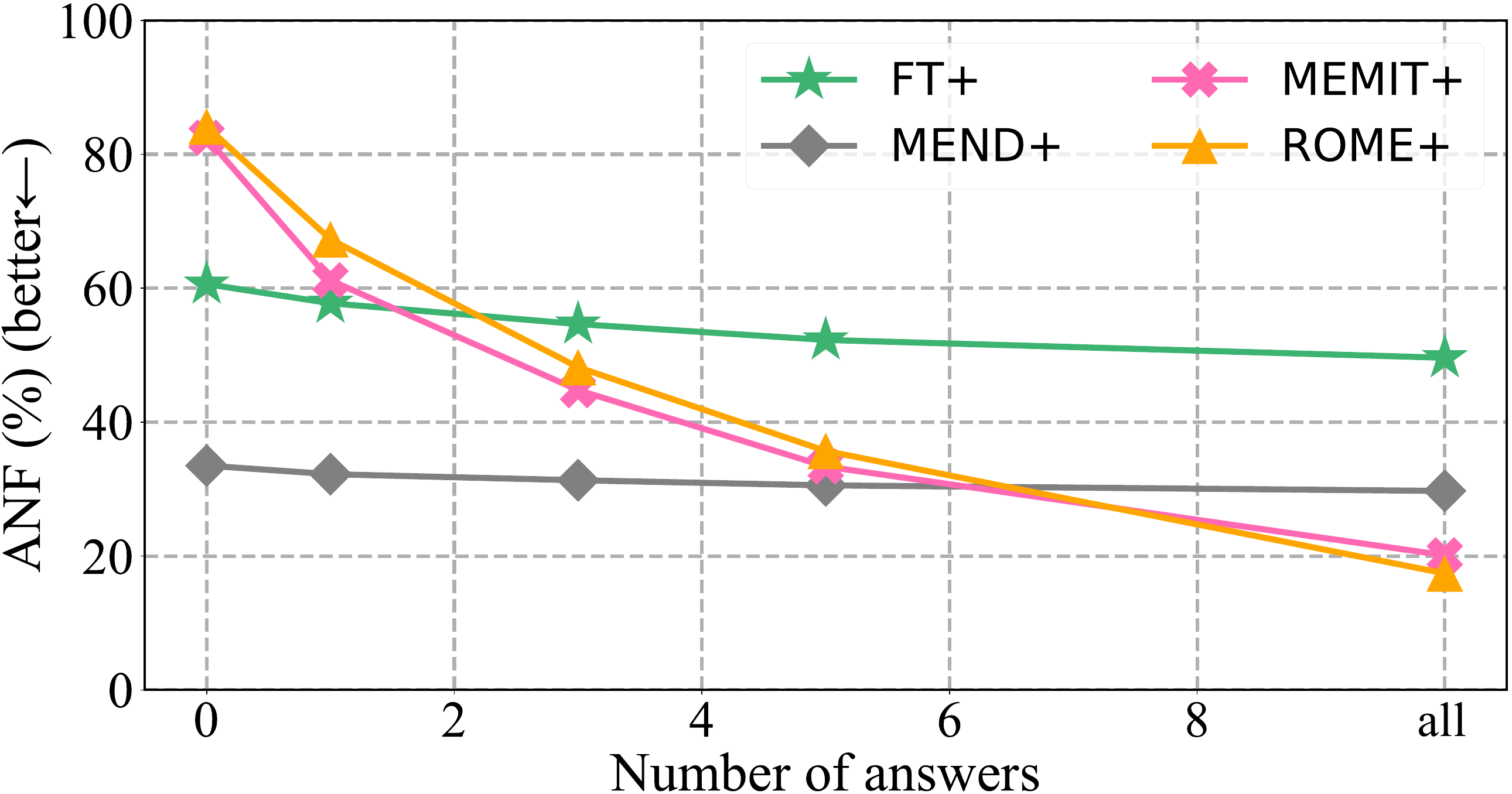}}
  \vspace{-3mm}
  \caption{Both AFF and ANF in \emph{Hard setting} on PEAK-CF with LLaMA-2 
  across four editing
methods equipped with APP where $k$ original correct and false answers were used.
$k \in [0, 1, 3, 5, all]$.}
  \vspace{-4mm}
  \label{numbers}
\end{figure}

Figure~\ref{numbers} illustrates how the performance of APP changed with respect to different numbers of neighboring answers on PEAK-CF with LLaMA-2 edited by different methods.
To this end, we extended our evaluation and considered utilizing $k$ correct and false answers in APP, where $k \in [0, 1, 3, 5, all]$.
It can be seen from these results that the performance of all editing methods coupled with APP was significantly improved as the number of neighboring answers increased.
This trend shows the effectiveness of considering more neighboring answers to comprehensively characterize the neighboring perturbations.
It also shows that both AFF and ANF can be significantly improved even with fewer answers, showcasing the practicality of the proposed APP method.

\section{Conclusion \& Limitation}
This study investigates the neighboring perturbations of knowledge editing on LLMs.
A metric of \emph{additivity} is introduced and a benchmark dubbed as PEAK is constructed for assessing the degree of perturbations in neighboring knowledge.
A plug-and-play framework APP is proposed to mitigate perturbations by minimizing the probability disruptions during knowledge appending. 
Comprehensive experiments on various knowledge editing methods and LLMs reveal that they inevitably perturb neighboring knowledge during editing, and the proposed APP method demonstrates its effectiveness in mitigating this perturbations to a certain extent.
In the future, we will explore expanding the scope of neighbor knowledge to more comprehensively characterize the neighboring perturbations of knowledge editing.

Besides, there are several limitations to our work.
First, regarding the experimental settings, the editing methods employed in this paper involve modifying the model parameters. But some editing methods that preserve model parameters remain to be explored. Second, this paper focuses on factual knowledge, and it would be beneficial to extend this approach to different types of knowledge, such as commonsense, logical, or spatial knowledge. Third, this paper designs the proposed additivity metric based on probability. However, the results generated by the actual model do not entirely depend on probability. Therefore, while this evaluation roughly reflects the degree of perturbation, it may deviate from real-world scenarios to some extent.

\section*{Acknowledgements}
We would like to express gratitude to the anonymous reviewers for their kind comments. 
This work is funded by the National Science and Technology Major Project (No. 2023ZD0121103).
Ningyu Zhang is supported by the National Natural Science Foundation of China (No. 62206246), Zhejiang Provincial Natural Science Foundation of China (No. LGG22F030011),  Yongjiang Talent Introduction Programme (2021A-156-G), CCF-Tencent Rhino-Bird Open Research Fund.

\section*{Impact Statement}
The research presented in this paper focuses on a critical aspect of LLMs – their susceptibility to unintended text generation due to false or outdated knowledge. In light of the resource-intensive nature of retraining LLMs, the study focuses on the emerging field of knowledge editing. 
It develops technologies to reduce perturbations to preserve factuality.
In terms of future societal consequences, 
this work contributes to mitigating the need for full retraining,  reducing computational resources and energy consumption. This aligns with sustainable AI and makes it more feasible to deploy LLMs in resource-constrained settings.
Besides, APP’s ability to perform targeted edits and mitigate neighboring perturbations can enhance LLMs' effectiveness in specialized domains, such as medical diagnostics, and legal analysis, where precise and updated knowledge is essential.
Additionally, editing methods for appending knowledge to LLMs could be misused to propagate misinformation. 
Furthermore, it could pose privacy risks if sensitive information is appended without proper safeguards.
In the future, some strategies should be implemented to reduce these risks.


\bibliography{base}

\begin{thebibliography}{42}
\providecommand{\natexlab}[1]{#1}
\providecommand{\url}[1]{\texttt{#1}}
\expandafter\ifx\csname urlstyle\endcsname\relax
  \providecommand{\doi}[1]{doi: #1}\else
  \providecommand{\doi}{doi: \begingroup \urlstyle{rm}\Url}\fi

\bibitem[Bau et~al.(2020)Bau, Liu, Wang, Zhu, and Torralba]{DBLP:conf/eccv/BauLWZT20}
Bau, D., Liu, S., Wang, T., Zhu, J., and Torralba, A.
\newblock Rewriting a deep generative model.
\newblock In Vedaldi, A., Bischof, H., Brox, T., and Frahm, J. (eds.), \emph{Computer Vision - {ECCV} 2020 - 16th European Conference, Glasgow, UK, August 23-28, 2020, Proceedings, Part {I}}, volume 12346 of \emph{Lecture Notes in Computer Science}, pp.\  351--369. Springer, 2020.
\newblock \doi{10.1007/978-3-030-58452-8\_21}.
\newblock URL \url{https://doi.org/10.1007/978-3-030-58452-8\_21}.

\bibitem[Cao et~al.(2021)Cao, Aziz, and Titov]{DBLP:conf/emnlp/CaoAT21}
Cao, N.~D., Aziz, W., and Titov, I.
\newblock Editing factual knowledge in language models.
\newblock In Moens, M., Huang, X., Specia, L., and Yih, S.~W. (eds.), \emph{Proceedings of the 2021 Conference on Empirical Methods in Natural Language Processing, {EMNLP} 2021, Virtual Event / Punta Cana, Dominican Republic, 7-11 November, 2021}, pp.\  6491--6506. Association for Computational Linguistics, 2021.
\newblock \doi{10.18653/v1/2021.emnlp-main.522}.
\newblock URL \url{https://doi.org/10.18653/v1/2021.emnlp-main.522}.

\bibitem[Cheng et~al.(2023)Cheng, Tian, Liu, Chen, Wang, Chen, and Zhang]{DBLP:conf/emnlp/0008TL0WC023}
Cheng, S., Tian, B., Liu, Q., Chen, X., Wang, Y., Chen, H., and Zhang, N.
\newblock Can we edit multimodal large language models?
\newblock In Bouamor, H., Pino, J., and Bali, K. (eds.), \emph{Proceedings of the 2023 Conference on Empirical Methods in Natural Language Processing, {EMNLP} 2023, Singapore, December 6-10, 2023}, pp.\  13877--13888. Association for Computational Linguistics, 2023.
\newblock URL \url{https://aclanthology.org/2023.emnlp-main.856}.

\bibitem[Cohen et~al.(2023)Cohen, Biran, Yoran, Globerson, and Geva]{DBLP:journals/corr/abs-2307-12976}
Cohen, R., Biran, E., Yoran, O., Globerson, A., and Geva, M.
\newblock Evaluating the ripple effects of knowledge editing in language models.
\newblock \emph{CoRR}, abs/2307.12976, 2023.
\newblock \doi{10.48550/ARXIV.2307.12976}.
\newblock URL \url{https://doi.org/10.48550/arXiv.2307.12976}.

\bibitem[Dai et~al.(2022)Dai, Dong, Hao, Sui, Chang, and Wei]{DBLP:conf/acl/DaiDHSCW22}
Dai, D., Dong, L., Hao, Y., Sui, Z., Chang, B., and Wei, F.
\newblock Knowledge neurons in pretrained transformers.
\newblock In Muresan, S., Nakov, P., and Villavicencio, A. (eds.), \emph{Proceedings of the 60th Annual Meeting of the Association for Computational Linguistics (Volume 1: Long Papers), {ACL} 2022, Dublin, Ireland, May 22-27, 2022}, pp.\  8493--8502. Association for Computational Linguistics, 2022.
\newblock \doi{10.18653/v1/2022.acl-long.581}.
\newblock URL \url{https://doi.org/10.18653/v1/2022.acl-long.581}.

\bibitem[Gandikota et~al.(2023)Gandikota, Materzynska, Fiotto{-}Kaufman, and Bau]{DBLP:journals/corr/abs-2303-07345}
Gandikota, R., Materzynska, J., Fiotto{-}Kaufman, J., and Bau, D.
\newblock Erasing concepts from diffusion models.
\newblock \emph{CoRR}, abs/2303.07345, 2023.
\newblock \doi{10.48550/ARXIV.2303.07345}.
\newblock URL \url{https://doi.org/10.48550/arXiv.2303.07345}.

\bibitem[Gentile \& Warmuth(1998)Gentile and Warmuth]{DBLP:conf/nips/GentileW98}
Gentile, C. and Warmuth, M.~K.
\newblock Linear hinge loss and average margin.
\newblock In Kearns, M.~J., Solla, S.~A., and Cohn, D.~A. (eds.), \emph{Advances in Neural Information Processing Systems 11, {[NIPS} Conference, Denver, Colorado, USA, November 30 - December 5, 1998]}, pp.\  225--231. The {MIT} Press, 1998.

\bibitem[Geva et~al.(2021)Geva, Schuster, Berant, and Levy]{DBLP:conf/emnlp/GevaSBL21}
Geva, M., Schuster, R., Berant, J., and Levy, O.
\newblock Transformer feed-forward layers are key-value memories.
\newblock In Moens, M., Huang, X., Specia, L., and Yih, S.~W. (eds.), \emph{Proceedings of the 2021 Conference on Empirical Methods in Natural Language Processing, {EMNLP} 2021, Virtual Event / Punta Cana, Dominican Republic, 7-11 November, 2021}, pp.\  5484--5495. Association for Computational Linguistics, 2021.
\newblock \doi{10.18653/v1/2021.emnlp-main.446}.
\newblock URL \url{https://doi.org/10.18653/v1/2021.emnlp-main.446}.

\bibitem[Geva et~al.(2022)Geva, Caciularu, Wang, and Goldberg]{DBLP:conf/emnlp/GevaCWG22}
Geva, M., Caciularu, A., Wang, K.~R., and Goldberg, Y.
\newblock Transformer feed-forward layers build predictions by promoting concepts in the vocabulary space.
\newblock In Goldberg, Y., Kozareva, Z., and Zhang, Y. (eds.), \emph{Proceedings of the 2022 Conference on Empirical Methods in Natural Language Processing, {EMNLP} 2022, Abu Dhabi, United Arab Emirates, December 7-11, 2022}, pp.\  30--45. Association for Computational Linguistics, 2022.
\newblock \doi{10.18653/v1/2022.emnlp-main.3}.
\newblock URL \url{https://doi.org/10.18653/v1/2022.emnlp-main.3}.

\bibitem[Gu et~al.(2024)Gu, Xu, Ma, Lu, Ling, Chang, and Peng]{DBLP:journals/corr/abs-2401-04700}
Gu, J., Xu, H., Ma, J., Lu, P., Ling, Z., Chang, K., and Peng, N.
\newblock Model editing can hurt general abilities of large language models.
\newblock \emph{CoRR}, abs/2401.04700, 2024.
\newblock \doi{10.48550/ARXIV.2401.04700}.
\newblock URL \url{https://doi.org/10.48550/arXiv.2401.04700}.

\bibitem[Huang et~al.(2023)Huang, Yu, Ma, Zhong, Feng, Wang, Chen, Peng, Feng, Qin, and Liu]{DBLP:journals/corr/abs-2311-05232}
Huang, L., Yu, W., Ma, W., Zhong, W., Feng, Z., Wang, H., Chen, Q., Peng, W., Feng, X., Qin, B., and Liu, T.
\newblock A survey on hallucination in large language models: Principles, taxonomy, challenges, and open questions.
\newblock \emph{CoRR}, abs/2311.05232, 2023.
\newblock \doi{10.48550/ARXIV.2311.05232}.
\newblock URL \url{https://doi.org/10.48550/arXiv.2311.05232}.

\bibitem[Ji et~al.(2023)Ji, Lee, Frieske, Yu, Su, Xu, Ishii, Bang, Madotto, and Fung]{DBLP:journals/csur/JiLFYSXIBMF23}
Ji, Z., Lee, N., Frieske, R., Yu, T., Su, D., Xu, Y., Ishii, E., Bang, Y., Madotto, A., and Fung, P.
\newblock Survey of hallucination in natural language generation.
\newblock \emph{{ACM} Comput. Surv.}, 55\penalty0 (12):\penalty0 248:1--248:38, 2023.
\newblock \doi{10.1145/3571730}.
\newblock URL \url{https://doi.org/10.1145/3571730}.

\bibitem[Li et~al.(2023)Li, Zhang, Yao, Wang, Chen, and Chen]{DBLP:journals/corr/abs-2310-02129}
Li, Z., Zhang, N., Yao, Y., Wang, M., Chen, X., and Chen, H.
\newblock Unveiling the pitfalls of knowledge editing for large language models.
\newblock \emph{CoRR}, abs/2310.02129, 2023.
\newblock \doi{10.48550/ARXIV.2310.02129}.
\newblock URL \url{https://doi.org/10.48550/arXiv.2310.02129}.

\bibitem[Ma et~al.(2022)Ma, Chen, Gu, Ling, Guo, Liu, Chen, and Liu]{DBLP:conf/emnlp/MaCGLGL0L22}
Ma, J., Chen, B., Gu, J., Ling, Z., Guo, W., Liu, Q., Chen, Z., and Liu, C.
\newblock Wider {\&} closer: Mixture of short-channel distillers for zero-shot cross-lingual named entity recognition.
\newblock In Goldberg, Y., Kozareva, Z., and Zhang, Y. (eds.), \emph{Proceedings of the 2022 Conference on Empirical Methods in Natural Language Processing, {EMNLP} 2022, Abu Dhabi, United Arab Emirates, December 7-11, 2022}, pp.\  5171--5183. Association for Computational Linguistics, 2022.
\newblock \doi{10.18653/v1/2022.emnlp-main.345}.
\newblock URL \url{https://doi.org/10.18653/v1/2022.emnlp-main.345}.

\bibitem[Ma et~al.(2023)Ma, Gu, Ling, Liu, and Liu]{DBLP:journals/corr/abs-2310-10322}
Ma, J., Gu, J., Ling, Z., Liu, Q., and Liu, C.
\newblock Untying the reversal curse via bidirectional language model editing.
\newblock \emph{CoRR}, abs/2310.10322, 2023.
\newblock \doi{10.48550/ARXIV.2310.10322}.
\newblock URL \url{https://doi.org/10.48550/arXiv.2310.10322}.

\bibitem[Mahdisoltani et~al.(2015)Mahdisoltani, Biega, and Suchanek]{DBLP:conf/cidr/MahdisoltaniBS15}
Mahdisoltani, F., Biega, J., and Suchanek, F.~M.
\newblock {YAGO3:} {A} knowledge base from multilingual wikipedias.
\newblock In \emph{Seventh Biennial Conference on Innovative Data Systems Research, {CIDR} 2015, Asilomar, CA, USA, January 4-7, 2015, Online Proceedings}. www.cidrdb.org, 2015.
\newblock URL \url{http://cidrdb.org/cidr2015/Papers/CIDR15\_Paper1.pdf}.

\bibitem[Mao et~al.(2023)Mao, Zhang, Wang, Wang, Yao, Jiang, Xie, Huang, and Chen]{DBLP:journals/corr/abs-2310-02168}
Mao, S., Zhang, N., Wang, X., Wang, M., Yao, Y., Jiang, Y., Xie, P., Huang, F., and Chen, H.
\newblock Editing personality for llms.
\newblock \emph{CoRR}, abs/2310.02168, 2023.
\newblock \doi{10.48550/ARXIV.2310.02168}.
\newblock URL \url{https://doi.org/10.48550/arXiv.2310.02168}.

\bibitem[Meng et~al.(2022)Meng, Bau, Andonian, and Belinkov]{DBLP:conf/nips/MengBAB22}
Meng, K., Bau, D., Andonian, A., and Belinkov, Y.
\newblock Locating and editing factual associations in {GPT}.
\newblock In \emph{NeurIPS}, 2022.
\newblock URL \url{https://arxiv.org/abs/2202.05262}.

\bibitem[Meng et~al.(2023)Meng, Sharma, Andonian, Belinkov, and Bau]{DBLP:conf/iclr/MengSABB23}
Meng, K., Sharma, A.~S., Andonian, A.~J., Belinkov, Y., and Bau, D.
\newblock Mass-editing memory in a transformer.
\newblock In \emph{The Eleventh International Conference on Learning Representations, {ICLR} 2023, Kigali, Rwanda, May 1-5, 2023}. OpenReview.net, 2023.
\newblock URL \url{https://openreview.net/pdf?id=MkbcAHIYgyS}.

\bibitem[Mitchell et~al.(2022{\natexlab{a}})Mitchell, Lin, Bosselut, Finn, and Manning]{DBLP:conf/iclr/MitchellLBFM22}
Mitchell, E., Lin, C., Bosselut, A., Finn, C., and Manning, C.~D.
\newblock Fast model editing at scale.
\newblock In \emph{The Tenth International Conference on Learning Representations, {ICLR} 2022, Virtual Event, April 25-29, 2022}. OpenReview.net, 2022{\natexlab{a}}.
\newblock URL \url{https://openreview.net/forum?id=0DcZxeWfOPt}.

\bibitem[Mitchell et~al.(2022{\natexlab{b}})Mitchell, Lin, Bosselut, Manning, and Finn]{DBLP:conf/icml/MitchellLBMF22}
Mitchell, E., Lin, C., Bosselut, A., Manning, C.~D., and Finn, C.
\newblock Memory-based model editing at scale.
\newblock In Chaudhuri, K., Jegelka, S., Song, L., Szepesv{\'{a}}ri, C., Niu, G., and Sabato, S. (eds.), \emph{International Conference on Machine Learning, {ICML} 2022, 17-23 July 2022, Baltimore, Maryland, {USA}}, volume 162 of \emph{Proceedings of Machine Learning Research}, pp.\  15817--15831. {PMLR}, 2022{\natexlab{b}}.
\newblock URL \url{https://proceedings.mlr.press/v162/mitchell22a.html}.

\bibitem[OpenAI(2023)]{DBLP:journals/corr/abs-2303-08774}
OpenAI.
\newblock {GPT-4} technical report.
\newblock \emph{CoRR}, abs/2303.08774, 2023.
\newblock \doi{10.48550/ARXIV.2303.08774}.
\newblock URL \url{https://doi.org/10.48550/arXiv.2303.08774}.

\bibitem[Ouyang et~al.(2022)Ouyang, Wu, Jiang, Almeida, Wainwright, Mishkin, Zhang, Agarwal, Slama, Ray, Schulman, Hilton, Kelton, Miller, Simens, Askell, Welinder, Christiano, Leike, and Lowe]{DBLP:conf/nips/Ouyang0JAWMZASR22}
Ouyang, L., Wu, J., Jiang, X., Almeida, D., Wainwright, C.~L., Mishkin, P., Zhang, C., Agarwal, S., Slama, K., Ray, A., Schulman, J., Hilton, J., Kelton, F., Miller, L., Simens, M., Askell, A., Welinder, P., Christiano, P.~F., Leike, J., and Lowe, R.
\newblock Training language models to follow instructions with human feedback.
\newblock In \emph{NeurIPS}, 2022.
\newblock URL \url{https://arxiv.org/abs/2203.02155}.

\bibitem[Pearl(2001)]{DBLP:conf/uai/Pearl01}
Pearl, J.
\newblock Direct and indirect effects.
\newblock In Breese, J.~S. and Koller, D. (eds.), \emph{{UAI} '01: Proceedings of the 17th Conference in Uncertainty in Artificial Intelligence, University of Washington, Seattle, Washington, USA, August 2-5, 2001}, pp.\  411--420. Morgan Kaufmann, 2001.
\newblock URL \url{https://dslpitt.org/uai/displayArticleDetails.jsp?mmnu=1\&smnu=2\&article\_id=126\&proceeding\_id=17}.

\bibitem[Peng et~al.(2023)Peng, Galley, He, Cheng, Xie, Hu, Huang, Liden, Yu, Chen, and Gao]{DBLP:journals/corr/abs-2302-12813}
Peng, B., Galley, M., He, P., Cheng, H., Xie, Y., Hu, Y., Huang, Q., Liden, L., Yu, Z., Chen, W., and Gao, J.
\newblock Check your facts and try again: Improving large language models with external knowledge and automated feedback.
\newblock \emph{CoRR}, abs/2302.12813, 2023.
\newblock \doi{10.48550/arXiv.2302.12813}.
\newblock URL \url{https://doi.org/10.48550/arXiv.2302.12813}.

\bibitem[Radford et~al.(2019)Radford, Wu, Child, Luan, Amodei, Sutskever, et~al.]{radford2019language}
Radford, A., Wu, J., Child, R., Luan, D., Amodei, D., Sutskever, I., et~al.
\newblock Language models are unsupervised multitask learners.
\newblock \emph{OpenAI blog}, 1\penalty0 (8):\penalty0 9, 2019.

\bibitem[Sinitsin et~al.(2020)Sinitsin, Plokhotnyuk, Pyrkin, Popov, and Babenko]{DBLP:conf/iclr/SinitsinPPPB20}
Sinitsin, A., Plokhotnyuk, V., Pyrkin, D.~V., Popov, S., and Babenko, A.
\newblock Editable neural networks.
\newblock In \emph{8th International Conference on Learning Representations, {ICLR} 2020, Addis Ababa, Ethiopia, April 26-30, 2020}. OpenReview.net, 2020.
\newblock URL \url{https://openreview.net/forum?id=HJedXaEtvS}.

\bibitem[Touvron et~al.(2023)Touvron, Martin, Stone, Albert, Almahairi, Babaei, Bashlykov, Batra, Bhargava, Bhosale, Bikel, Blecher, Canton{-}Ferrer, Chen, Cucurull, Esiobu, Fernandes, Fu, Fu, Fuller, Gao, Goswami, Goyal, Hartshorn, Hosseini, Hou, Inan, Kardas, Kerkez, Khabsa, Kloumann, Korenev, Koura, Lachaux, Lavril, Lee, Liskovich, Lu, et~al.]{DBLP:journals/corr/abs-2307-09288}
Touvron, H., Martin, L., Stone, K., Albert, P., Almahairi, A., Babaei, Y., Bashlykov, N., Batra, S., Bhargava, P., Bhosale, S., Bikel, D., Blecher, L., Canton{-}Ferrer, C., Chen, M., Cucurull, G., Esiobu, D., Fernandes, J., Fu, J., Fu, W., Fuller, B., Gao, C., Goswami, V., Goyal, N., Hartshorn, A., Hosseini, S., Hou, R., Inan, H., Kardas, M., Kerkez, V., Khabsa, M., Kloumann, I., Korenev, A., Koura, P.~S., Lachaux, M., Lavril, T., Lee, J., Liskovich, D., Lu, Y., et~al.
\newblock Llama 2: Open foundation and fine-tuned chat models.
\newblock \emph{CoRR}, abs/2307.09288, 2023.
\newblock \doi{10.48550/arXiv.2307.09288}.
\newblock URL \url{https://doi.org/10.48550/arXiv.2307.09288}.

\bibitem[Vig et~al.(2020)Vig, Gehrmann, Belinkov, Qian, Nevo, Singer, and Shieber]{DBLP:conf/nips/VigGBQNSS20}
Vig, J., Gehrmann, S., Belinkov, Y., Qian, S., Nevo, D., Singer, Y., and Shieber, S.~M.
\newblock Investigating gender bias in language models using causal mediation analysis.
\newblock In Larochelle, H., Ranzato, M., Hadsell, R., Balcan, M., and Lin, H. (eds.), \emph{Advances in Neural Information Processing Systems 33: Annual Conference on Neural Information Processing Systems 2020, NeurIPS 2020, December 6-12, 2020, virtual}, 2020.

\bibitem[Vrandecic \& Kr{\"{o}}tzsch(2014)Vrandecic and Kr{\"{o}}tzsch]{DBLP:journals/cacm/VrandecicK14}
Vrandecic, D. and Kr{\"{o}}tzsch, M.
\newblock Wikidata: a free collaborative knowledgebase.
\newblock \emph{Commun. {ACM}}, 57\penalty0 (10):\penalty0 78--85, 2014.
\newblock \doi{10.1145/2629489}.
\newblock URL \url{https://doi.org/10.1145/2629489}.

\bibitem[Wang \& Komatsuzaki(2021)Wang and Komatsuzaki]{wang2021gpt}
Wang, B. and Komatsuzaki, A.
\newblock Gpt-j-6b: A 6 billion parameter autoregressive language model, 2021.

\bibitem[Wang et~al.(2023)Wang, Zhang, Xie, Yao, Tian, Wang, Xi, Cheng, Liu, Zheng, and Chen]{DBLP:journals/corr/abs-2308-07269}
Wang, P., Zhang, N., Xie, X., Yao, Y., Tian, B., Wang, M., Xi, Z., Cheng, S., Liu, K., Zheng, G., and Chen, H.
\newblock Easyedit: An easy-to-use knowledge editing framework for large language models.
\newblock \emph{CoRR}, abs/2308.07269, 2023.
\newblock \doi{10.48550/arXiv.2308.07269}.
\newblock URL \url{https://doi.org/10.48550/arXiv.2308.07269}.

\bibitem[Wang et~al.(2024)Wang, Chen, Peng, and Chang]{wang2024deepedit}
Wang, Y., Chen, M., Peng, N., and Chang, K.-W.
\newblock Deepedit: Knowledge editing as decoding with constraints.
\newblock \emph{arXiv preprint arXiv:2401.10471}, 2024.

\bibitem[Wu et~al.(2023)Wu, Li, Xu, Dong, Wu, Bian, and Xiong]{DBLP:conf/emnlp/WuLXDW0X23}
Wu, X., Li, J., Xu, M., Dong, W., Wu, S., Bian, C., and Xiong, D.
\newblock {DEPN:} detecting and editing privacy neurons in pretrained language models.
\newblock In Bouamor, H., Pino, J., and Bali, K. (eds.), \emph{Proceedings of the 2023 Conference on Empirical Methods in Natural Language Processing, {EMNLP} 2023, Singapore, December 6-10, 2023}, pp.\  2875--2886. Association for Computational Linguistics, 2023.
\newblock URL \url{https://aclanthology.org/2023.emnlp-main.174}.

\bibitem[Wu \& Liu(2007)Wu and Liu]{wu2007robust}
Wu, Y. and Liu, Y.
\newblock Robust truncated hinge loss support vector machines.
\newblock \emph{Journal of the American Statistical Association}, 102\penalty0 (479):\penalty0 974--983, 2007.

\bibitem[Yao et~al.(2023)Yao, Wang, Tian, Cheng, Li, Deng, Chen, and Zhang]{DBLP:conf/emnlp/YaoWT0LDC023}
Yao, Y., Wang, P., Tian, B., Cheng, S., Li, Z., Deng, S., Chen, H., and Zhang, N.
\newblock Editing large language models: Problems, methods, and opportunities.
\newblock In Bouamor, H., Pino, J., and Bali, K. (eds.), \emph{Proceedings of the 2023 Conference on Empirical Methods in Natural Language Processing, {EMNLP} 2023, Singapore, December 6-10, 2023}, pp.\  10222--10240. Association for Computational Linguistics, 2023.
\newblock URL \url{https://aclanthology.org/2023.emnlp-main.632}.

\bibitem[Yin et~al.(2023)Yin, Jiang, Yang, and Wan]{DBLP:journals/corr/abs-2312-05497}
Yin, X., Jiang, J., Yang, L., and Wan, X.
\newblock History matters: Temporal knowledge editing in large language model.
\newblock \emph{CoRR}, abs/2312.05497, 2023.
\newblock \doi{10.48550/ARXIV.2312.05497}.
\newblock URL \url{https://doi.org/10.48550/arXiv.2312.05497}.

\bibitem[Zhang et~al.(2023{\natexlab{a}})Zhang, Haddow, and Birch]{DBLP:conf/icml/0006HB23}
Zhang, B., Haddow, B., and Birch, A.
\newblock Prompting large language model for machine translation: {A} case study.
\newblock In Krause, A., Brunskill, E., Cho, K., Engelhardt, B., Sabato, S., and Scarlett, J. (eds.), \emph{International Conference on Machine Learning, {ICML} 2023, 23-29 July 2023, Honolulu, Hawaii, {USA}}, volume 202 of \emph{Proceedings of Machine Learning Research}, pp.\  41092--41110. {PMLR}, 2023{\natexlab{a}}.
\newblock URL \url{https://proceedings.mlr.press/v202/zhang23m.html}.

\bibitem[Zhang et~al.(2024)Zhang, Yao, Tian, Wang, Deng, Wang, Xi, Mao, Zhang, Ni, et~al.]{zhang2024comprehensive}
Zhang, N., Yao, Y., Tian, B., Wang, P., Deng, S., Wang, M., Xi, Z., Mao, S., Zhang, J., Ni, Y., et~al.
\newblock A comprehensive study of knowledge editing for large language models.
\newblock \emph{arXiv preprint arXiv:2401.01286}, 2024.

\bibitem[Zhang et~al.(2023{\natexlab{b}})Zhang, Li, Cui, Cai, Liu, Fu, Huang, Zhao, Zhang, Chen, Wang, Luu, Bi, Shi, and Shi]{DBLP:journals/corr/abs-2309-01219}
Zhang, Y., Li, Y., Cui, L., Cai, D., Liu, L., Fu, T., Huang, X., Zhao, E., Zhang, Y., Chen, Y., Wang, L., Luu, A.~T., Bi, W., Shi, F., and Shi, S.
\newblock Siren's song in the {AI} ocean: {A} survey on hallucination in large language models.
\newblock \emph{CoRR}, abs/2309.01219, 2023{\natexlab{b}}.
\newblock \doi{10.48550/arXiv.2309.01219}.
\newblock URL \url{https://doi.org/10.48550/arXiv.2309.01219}.

\bibitem[Zhong et~al.(2023)Zhong, Wu, Manning, Potts, and Chen]{DBLP:conf/emnlp/ZhongWMPC23}
Zhong, Z., Wu, Z., Manning, C.~D., Potts, C., and Chen, D.
\newblock Mquake: Assessing knowledge editing in language models via multi-hop questions.
\newblock In Bouamor, H., Pino, J., and Bali, K. (eds.), \emph{Proceedings of the 2023 Conference on Empirical Methods in Natural Language Processing, {EMNLP} 2023, Singapore, December 6-10, 2023}, pp.\  15686--15702. Association for Computational Linguistics, 2023.
\newblock URL \url{https://aclanthology.org/2023.emnlp-main.971}.

\bibitem[Zhu et~al.(2020)Zhu, Rawat, Zaheer, Bhojanapalli, Li, Yu, and Kumar]{DBLP:journals/corr/abs-2012-00363}
Zhu, C., Rawat, A.~S., Zaheer, M., Bhojanapalli, S., Li, D., Yu, F.~X., and Kumar, S.
\newblock Modifying memories in transformer models.
\newblock \emph{CoRR}, abs/2012.00363, 2020.
\newblock URL \url{https://arxiv.org/abs/2012.00363}.

\end{thebibliography}
\bibliographystyle{icml2024}

\newpage
\appendix
\onecolumn

\section{Datasets}

\subsection{Construction of False Answers} \label{append-cons}
For the \emph{additivity} proposed in this paper, 
given an edit $\mathcal{E}=\left\{s, r, O, o^{*}, p\right\}$ where $t_r(s)=p$,
both the original correct answers and sampled false answers are utilized.
The false answers were sampled in two settings \emph{Hard} and \emph{Random}.
For the \emph{Hard} setting, some objects establish direct relations with the new object $o^*$ are selected.
Specifically, denote the triples of Wikidata as $W(r)= \{(s, r, o )\}$, we firstly seek entities in $W(r)$ which have a link $r^\prime$ with $o^*$ (could be the form of $(o^*, r^\prime, *)$, $(*, r^\prime, o^*)$), then we collect them and choose the entities has same entity type with $o^*$.
For example, if $o^*$ is a person, then the selected entities are persons, too.
Finally, we sample some selected entities as the false answers.
For the \emph{Random} setting, we seek entities do not have links with $o^*$, while other parts are the same.
In this way, the false answers in the \emph{Hard} setting are more semantically close to the new appended object, while false answers in the \emph{Random} setting are semantically distant from the new object.

\subsection{Relations and Templates of Datasets} \label{append-temple}
Table~\ref{handwrittencf} and ~\ref{handwrittent} show some examples of relations and their templates.

\begin{table*}[h]
\setlength{\abovecaptionskip}{0.1cm} 
\caption{Part of templates $T(r)$ for relation $r$ in PEAK-CF. Actually, there are several templates for each relation. Here we only display one or two templates for each relation.} \label{handwrittencf}
\centering
\setlength{\tabcolsep}{3mm}{
\begin{tabular}{llcc}
\toprule
& Relation (r) & $T(r)$ \\
\midrule
& producer  & ``\{\} is the producer of what products?'', ``\{\} has produced many products such as' \\
& official language &``\{\}, which is the official language of'', ``\{\} has been the official language of'' \\
& illustration  & ``The illustrator \{\} has created illustrations'' \\
& written  & ``\{\}, who has written some books like'', ``Which books have been written by \{\}?'' \\
\bottomrule

\end{tabular}}

\end{table*}

\begin{table*}[h]
\setlength{\abovecaptionskip}{0.1cm} 
\caption{Part of templates $T(r)$ for relation $r$ in PEAK-T.} \label{handwrittent}
\centering
\setlength{\tabcolsep}{3mm}{
\begin{tabular}{llcc}
\toprule
& Relation (r) & $T(r)$ \\
\midrule
& plays for  & ``For so long, \{\} has been at which clubs?'', `` What clubs have hired \{\} as a player?''\\
& created &``\{\} was the individual responsible for creating'' \\
& has won prize  & ``\{\} was the recipient of the prize'',``What prizes have been gained by \{\}?'' \\

\bottomrule

\end{tabular}}

\end{table*}

\section{Experimental Setup}

\subsection{Baseline Methods} \label{append_baselines}
Five popular model editing methods were selected as baselines including:
\begin{itemize}
\item[$\bullet$] \textbf{FT}~\cite{DBLP:journals/corr/abs-2012-00363}: this method simply performed gradient descent on the edits to update model parameters.
It fine-tuned
one layer in the model with a norm constraint
on weight changes to prevent overfitting.
Since the original authors did not publish their code, we followed \citet{DBLP:conf/nips/MengBAB22}
re-implementation in their study.
\item[$\bullet$] \textbf{KN}~\cite{DBLP:conf/acl/DaiDHSCW22}\footnote{https://github.com/EleutherAI/knowledge-neurons.}: it first selected neurons that were associated with knowledge expression via gradient-based attributions,
and then modified MLP layer at the rows corresponding to those neurons by adding scaled embedding vectors.
\item[$\bullet$] \textbf{MEND}~\cite{DBLP:conf/iclr/MitchellLBFM22}\footnote{https://github.com/eric-mitchell/mend}: it learned a hypernetwork to produce weight updates by decomposing
the fine-tuning gradients into rank-1 form.
\item[$\bullet$] \textbf{ROME}~\cite{DBLP:conf/nips/MengBAB22}\footnote{https://github.com/kmeng01/rome}: it first localized
the factual knowledge at a specific layer in the
transformer MLP modules, and then updated the knowledge by directly writing
new key-value pairs in the MLP module.
\item[$\bullet$]
\textbf{MEMIT}~\cite{DBLP:conf/iclr/MengSABB23}\footnote{https://github.com/kmeng01/memit}: it extended ROME
to edit a large set of facts and updated a sequence of MLP layers to update knowledge.
\end{itemize}

The ability of these methods were assessed based on
EasyEdit~\cite{DBLP:journals/corr/abs-2308-07269}, an easy-to-use knowledge
editing framework which integrates the released codes and hyperparameters from previous methods.

\subsection{Hyperparameters for APP}\label{append-hyper}
To set the hyperparameters, we additionally created a small validation set.
Table~\ref{hyperparameters} shows the details of hyperparameters set for different LLMs.
Besides, the margin $\mathcal{M}$ is set to 2.

\begin{table*}[ht]
\vspace{-2mm}
\footnotesize
\centering
\setlength{\abovecaptionskip}{0.1cm} 
\caption{The hyperparameters for APP on each method on each model.} \label{hyperparameters}
\renewcommand\arraystretch{1}
\setlength{\tabcolsep}{3.5mm}{
\begin{tabular}{lccc|ccc|ccc}
\toprule
\multirow{2}{*}{\textbf{ Editor }} 
& \multicolumn{3}{c}{\textbf { GPT-2 XL (1.5B) }} 
& \multicolumn{3}{c}{\textbf { GPT-J (6B)  }} 
& \multicolumn{3}{c}{\textbf { LLaMA-2 (7B) }} 
 \\
\cmidrule(r){2-4}
\cmidrule(r){5-7}
\cmidrule(r){8-10}
& $\alpha$ & $\beta$ & $\gamma$ & $\alpha$ & $\beta$ & $\gamma$ & $\alpha$ & $\beta$ & $\gamma$   \\
 \midrule[0.5pt]
 \text {FT+APP} &0.2 &0.5 &0.2 &0.2 &0.5 &0.2 &0.2 &0.5 &0.2
 \\
  \text {MEND+APP} & 0.5 &0 &0 &0.5 &0 &0 &0.3 &0 &0

  \\
  \text {MEMIT+APP} &0.05 &0.05 &0.05 &0.1 &0.3 &0.2 &0.6 &1.0 &0.3

  \\
  \text {ROME+APP} &0.2 &0.2 &0.1 &0.2 &0.2 &0.1 &0.5 &0.8 &0.1
\\
\bottomrule
\end{tabular}}
\vspace{-2mm}
\end{table*}

\subsection{Examples of previous metrics} \label{append-metric}
Here we provide an example to show what each previous metric means.

\paragraph{Efficacy} Given an editing fact ($s$ = Apple, $r$ = products, $O$ = \{AirPods 3, MacBook Air,..., iPhone 14\}, $o^{*}$=iPhone 15),
if the edited model $\mathcal{F}^{*}$ assigns a higher probability to the new object $o^{*}$ than the minimum probability in $O$\footnote{Previous works care about whether the probability surpasses the original $o$. Here $O$ is a set, so we made slight changes.} under the corresponding question $t_r(s)$ ``What are the products of Apple?'', then this new knowledge is appended effectively. 

\paragraph{Generalization} 
The edited model is considered to have generalized successfully if it can recall the new knowledge with paraphrased prompt like ``What items does Apple produce?''.
That is, to determine whether the probability assigned by the edited model $\mathcal{F}^{*}$ to the new object $o^{*}$ is higher than the minimum probability in $O$ under the paraphrased prompt.

\paragraph{Locality} The edited model should remain
unchanged in response to prompts that are irrelevant or outside the scope of editing. For example, the answer to the question ``Which company developed Windows?'' should still be ``Microsoft''.

\subsection{Results of LLaMA-2 (13B)} \label{append-13b}
The results of LLaMA-2 (13B) are shown in Table~\ref{13b}.
Compared with previous results, we can conclude that previous important conclusions are still valid, such as ``The performance of memorizing the new target knowledge is good'', ``larger LLMs suffer more serious perturbation'' and ``APP significantly mitigates the perturbation''.

\begin{table}[ht]
\footnotesize
\centering
\caption{Evaluation results (\%) on the PEAK-CF dataset with LLaMA-2 (13B).
}
\label{13b}
\setlength{\tabcolsep}{2mm}{
\begin{tabular}{lccccccc}
\toprule
\multirow{3}{*}{\textbf{ Editor }} 
& \multicolumn{7}{c}{\textbf { LLaMA-2 (13B) }} 

 \\
\cmidrule(r){2-8}
&\multicolumn{3}{c}{\textbf { Previous }} 
&\multicolumn{2}{c}{\textbf { Additivity (hard) }} 
&\multicolumn{2}{c}{\textbf { Additivity (ran) }} 
\\
\cmidrule(r){2-4}
\cmidrule(r){5-6}
\cmidrule(r){7-8}
& $\mathrm{ES} \uparrow$ & $\mathrm{GS} \uparrow$  & $\mathrm{LS} \uparrow$ & $\mathrm{AFF} \downarrow$ & $\mathrm{ANF} \downarrow$ & $\mathrm{AFF} \downarrow$ & $\mathrm{ANF} \downarrow$ \\
 \midrule[0.5pt]
 \text {FT}  &99.66	&84.46	&78.16	&81.30	&57.86	&73.45	&37.65

 \\
  \text {MEMIT} &99.66	&98.47	&93.50	&94.00	&85.92	&86.10	&56.18
  \\
  \text {ROME} &100.00	&97.90	&92.91	&93.09	&84.93	&83.46	&54.29
  \\

\hline\\[-2.0mm]\hline
\rule{0pt}{9pt} 

 \text {FT+APP}  &97.61	&81.07	&80.75	&71.04	&47.30	&61.22	&32.19

 \\
  \text {MEMIT+APP} &97.95	&92.26	&96.13	&51.03	&26.20	&44.92	&20.19
  \\
  \text {ROME+APP} &96.59	&91.41	&94.97	&42.08	&23.09	&35.25	&17.26
  \\
\bottomrule
\end{tabular}}
\end{table}

\subsection{Ablation study} \label{append-ablation}
\begin{figure}[t]
  \centering
  \subfigure{
  \includegraphics[width=7.09cm]{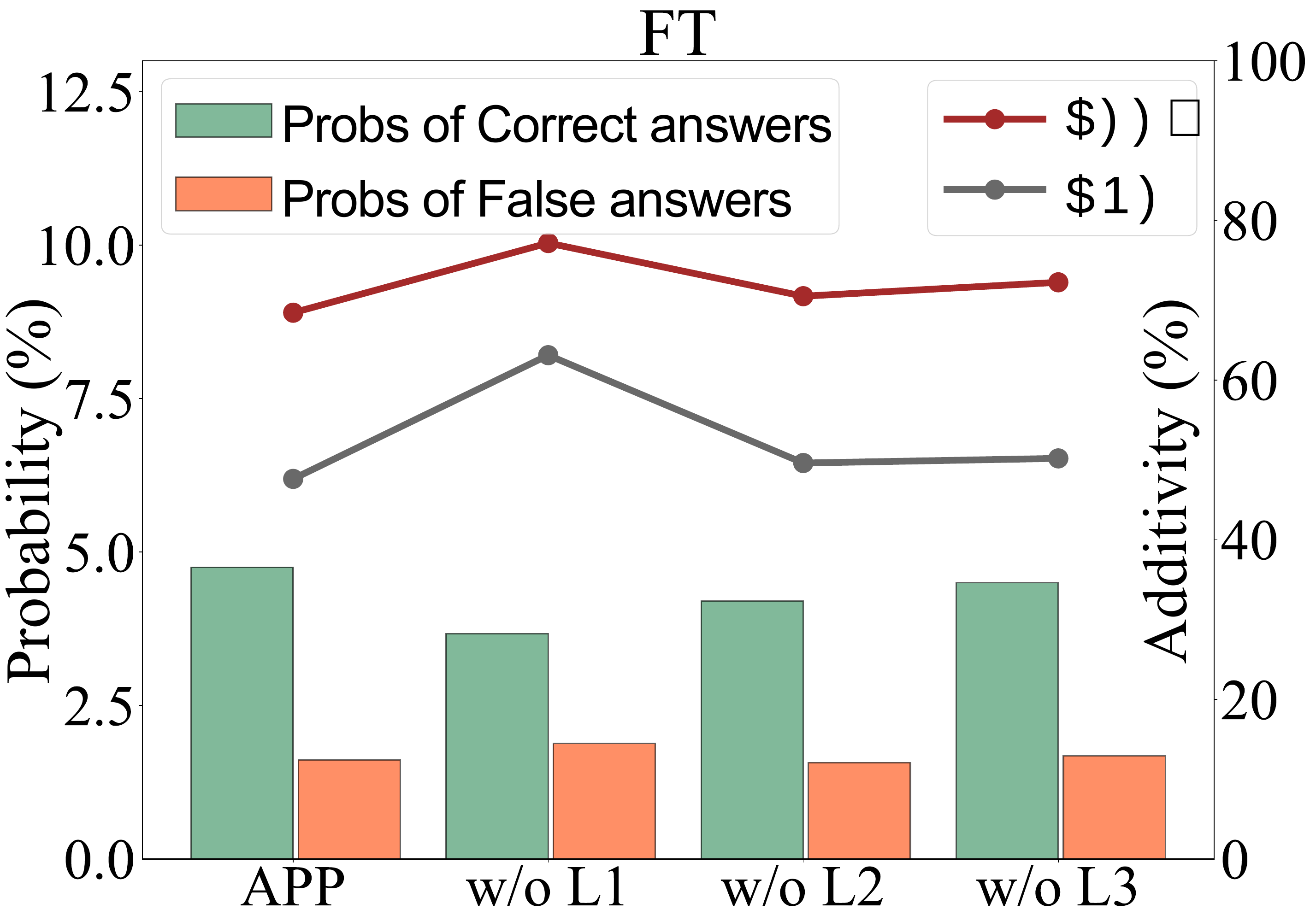}}
  \vspace{-4mm}
  \caption{ Ablation analysis of probability and additivity for APP.
  Results were conducted with LLaMA-2 on PEAK-CF dataset.}
  \label{ab-ft}
\end{figure}

Figure~\ref{ab-ft} shows other results of the ablation study in Section~\ref{sec-ablation}.
As MEND could only use $\mathcal{L}_1$ in APP, we only list the results of FT.

\section{Examples of how APP is coupled with existing methods}
\label{exampleapp}
Here we take the editing method ROME as an example.
\subsection{Rank-One Model Editing (ROME)}\label{ro}
ROME~\cite{DBLP:conf/nips/MengBAB22} applies causal mediation analysis~\cite{DBLP:conf/uai/Pearl01,DBLP:conf/nips/VigGBQNSS20} to locate the MLP modules that store facts.
An MLP module consists of two layers, where the first and second layers are denoted as $W_{f c}^{l}$ and $W_{proj}^{l}$ respectively.
$W_{proj}^{l}$ is considered as a linear associative memory and an editing area.
A new fact is represented as a key-value pair ($k_{*}$, $ v_{*}$)~\cite{DBLP:conf/emnlp/GevaSBL21}, which can be inserted into the MLP module of the model by solving a constrained least-squares problem~\cite{DBLP:conf/eccv/BauLWZT20}. 
All that remains is to choose the appropriate $k_{*}$ and $ v_{*}$.

Denote $L$ as an edited MLP layer, $\mathcal{F}$ as the unedited model, and $e=(s, r, o^{*})$ as the editing fact.
The key $k_{*}$ is obtained by calculating the average representation of the \textbf{last} token of the subject $s$ outputted by $W_{f c}^{L}$ as follows:
\vspace{-2mm}
\begin{equation}
        \begin{aligned}
k_{*}&=\frac{1}{N} \sum_{j=1}^{N} k\left(x_{j}+s\right) \text {,} \\ \text{where } k(x)&=\sigma\left(W_{f c}^{\left(L\right)} \gamma\left(a_{[x], i}^{\left(L\right)}+h_{[x], i}^{\left(L-1\right)}\right)\right) \text {. }
    \end{aligned}
\vspace{-2mm}
\end{equation}

Here, $x_{j}$ is the random sampled sequence and $i$ is the location of the last token of the subject in sentence $x$.
$h$ is the hidden states, $a$ is the attention, $\gamma$ is the layernorm, and $\sigma$ is the activation function.

The value $v_{*}$ encodes the new knowledge ($r$, $o^{*}$) as a property of $s$.
To calculate it, ROME sets $v_{*}=\operatorname{argmin}_{z} \mathcal{L}(z)$,  where the objective is denoted as:
\vspace{-2mm}
\begin{equation}
        \begin{aligned}
    \mathcal{L}_e (o^*, z) = -\log P_{\mathcal{F}\left(m_{i}^{\left(L\right)}:=z\right)}\left[o^{*} \mid p\right]+ 
    D_{\mathrm{KL}}\left(P_{\mathcal{F}\left(m_{i^{\prime}}^{\left(L\right)}:=z\right)}\left[x \mid p^{\prime}\right] \| P_{\mathcal{F}}\left[x \mid p^{\prime}\right]\right) \text {. }
    \end{aligned}\label{rome}
\end{equation}

\vspace{-2mm}
The first term in Eq.~(\ref{rome}) seeks a vector z that, when substituted as the output of the MLP \textbf{at the token $i$} (notated  $\mathcal{F}\left(m_{i}^{\left(L\right)}:=z\right)$ 
 and here $i$ is the end of the subject in prompt $p$), will cause the network to predict the target object  $o^{*}$  in response to the factual prompt  $p$ (e.g., $p$=Eiffel Tower is located in, $o^{*}$=London). 
 The second term minimizes the KL divergence~\cite{DBLP:conf/emnlp/MaCGLGL0L22} of predictions for the prompt  $p^{\prime}$  (of the form "  \{  subject  \}  is a") to the unchanged model, which helps preserve the model's understanding of the subject's essence ($i^{\prime}$ is the location of the last token of the subject in $p^{\prime}$).

Finally, the MLP weight $W_{proj}^{L}$ is updated with a rank-one update to insert the new fact:
\begin{equation}
        \begin{aligned}
\hat{W}=W+\Lambda\left(C^{-1} k_{*}\right)^{T} \text{,}
    \end{aligned} \label{finalequa}
\end{equation}
where $C$ is a constant by estimating the covariance of many existing keys. $\Lambda=\left(v_{*}-W k_{*}\right) /\left(C^{-1} k_{*}\right)^{T} k_{*}$, representing the residual error of the new key–value pair on the original memory matrix.

Readers can refer to \citet{DBLP:conf/nips/MengBAB22}
for more details of ROME.

\subsection{Apply APP to ROME}
In ROME, the editing objective is $\mathcal{L}_e(o^*, z)$, and the parameter to be optimized is $z$ for $v_{*}=\operatorname{argmin}_{z} \mathcal{L}(z)$.
Therefore, the proposed APP could be coupled with $\mathcal{L}_e(o^*, z)$ to optimize $z$.

On the one hand, the editing objective $\mathcal{L}_{1}(O, O_{h},z)$ designed to maintain a certain margin between the probabilities of original correct answers $O$ and those of false answers $O_{h}$ could be written as follows:
\begin{equation}
    \begin{aligned}
        \mathcal{L}_{1}(O, O_{h}, z) = \frac{1}{NM}\sum_{i=1}^{N}\sum_{j=1}^{M} max\{0, \: \mathcal{M}-\log P_{\mathcal{F}(m_{i}^{\left(L\right)}:=z)}(o_{i} \,|\, p)  + \log P_{\mathcal{F}(m_{i}^{\left(L\right)}:=z)} (o_{hj} \,|\, p)\},
    \end{aligned}
\end{equation}
$N$ and $M$ represent the number of elements in $O$ and $O_{h}$ respectively.
This equation seeks a $z$ that the log probabilities of correct answers are encouraged to be larger than those of false answers by at least a certain margin $\mathcal{M}$.

On the other hand, it involves ensuring
that the absolute probabilities of correct answers do not decrease
while simultaneously controlling that the probabilities of false answers do not increase during editing, which can be represented as two objectives $\mathcal{L}_{2}(O,z)$ and $\mathcal{L}_{3}(O_h, z)$ respectively as:
\begin{equation}
    \begin{aligned}
         &\mathcal{L}_{2}(O,z) = \frac{1}{N}\sum_{i=1}^{N}max\{0, \, \log P_{\mathcal{F}}(o_{i} \,|\, p) - \log P_{\mathcal{F}(m_{i}^{\left(L\right)}:=z)}(o_{i} \,|\, p)\}, 
    \end{aligned}
\vspace{-2mm}
\end{equation}

\begin{equation}
    \begin{aligned}
     &\mathcal{L}_{3}(O_s, z) = 
     \frac{1}{M}\sum_{i=1}^{M}max\{0, \, \log P_{\mathcal{F}(m_{i}^{\left(L\right)}:=z)}(o_{hi} \,|\, p)- \log P_{\mathcal{F}}(o_{hi} \,|\, p)\}. 
    \end{aligned}
\vspace{-2mm}
\end{equation}

$P_{\mathcal{F}}(o_{i} \,|\, p)$ refers the probability output by the unedited model.
$P_{\mathcal{F}(m_{i}^{\left(L\right)}:=z)}(o_{i} \,|\, p)$ refers the probability output by the model operated with the same operation in Eq.~(\ref{rome}).
These two equations respectively seek a $z$ that the log probabilities of correct answers are encouraged not to be decreased and the log probabilities of false answers not to be increased.
Finally, these proposed objectives are jointly optimized with the editing objective $\mathcal{L}_{e}(o^*, z)$ to obtain $v_{*}$:
\begin{equation}
    \begin{aligned}
            v_{*} = \mathop{\arg\min}\limits_{z} \mathcal{L}_{e}(o^*,z) + \alpha \mathcal{L}_{1}(O, O_{h},z) + \beta \mathcal{L}_{2}(O,z) +\gamma \mathcal{L}_{3}(O_{h},z).
    \end{aligned}
\end{equation}
Finally, the Eq.~(\ref{finalequa}) is utilized to update parameters in model.

\end{document}